\documentclass[lettersize,journal]{IEEEtran}
\usepackage{amsmath,amsfonts}
\usepackage{algorithmic}
\usepackage{algorithm}
\usepackage{array}
\usepackage[caption=false,font=normalsize,labelfont=sf,textfont=sf]{subfig}
\usepackage{textcomp}
\usepackage{stfloats}
\usepackage{url}
\usepackage{verbatim}
\usepackage{graphicx}
\usepackage{cite}
\usepackage{xcolor}
\usepackage[colorlinks=true, linkcolor=blue, citecolor=blue, urlcolor=black]{hyperref}
\usepackage{booktabs}
\usepackage{multirow}
\usepackage[normalem]{ulem}
\usepackage{tikz}
\usetikzlibrary{positioning}
\useunder{\uline}{\ul}{}
\begin{document}

\title{An Empirical Study of Sample Selection Strategies for Large Language Model Repair}

\author{Xuran Li, Jingyi Wang

\thanks{Xuran li and Jingyi Wang are with the Zhejiang University, Hangzhou, Zhejiang 310027, China. E-mail: xuranli1005@zju.edu.cn, wangjyee@zju.edu.cn.}}


\maketitle

\begin{abstract}
Large language models (LLMs) are increasingly deployed in real-world systems, yet they can produce toxic or biased outputs that undermine safety and trust. Post-hoc model repair provides a practical remedy, but the high cost of parameter updates motivates selective use of repair data. Despite extensive prior work on data selection for model training, it remains unclear which sampling criteria are most effective and efficient when applied specifically to behavioral repair of large generative models. Our study presents a systematic analysis of sample prioritization strategies for LLM repair. We evaluate five representative selection methods, including random sampling, K-Center, gradient-norm-based selection (GraNd), stratified coverage (CCS), and a Semantic-Aware Prioritized Sampling (SAPS) approach we proposed. Repair effectiveness and trade-offs are assessed through toxicity reduction, perplexity on WikiText-2 and LAMBADA, and three composite metrics: the Repair Proximity Score (RPS), the Overall Performance Score (OPS), and the Repair Efficiency Score (RES). Experimental results show that SAPS achieves the best balance between detoxification, utility preservation, and efficiency, delivering comparable or superior repair outcomes with substantially less data. Random sampling remains effective for large or robust models, while high-overhead methods such as CCS and GraNd provide limited benefit. The optimal data proportion depends on model scale and repair method, indicating that sample selection should be regarded as a tunable component of repair pipelines. Overall, these findings establish selection-based repair as an efficient and scalable paradigm for maintaining LLM reliability.
\end{abstract}

\begin{IEEEkeywords}
Large Language Models, Model Repair, Sample Selection, Behavioral Correction, Empirical Study.
\end{IEEEkeywords}

\section{Introduction}
\IEEEPARstart{L}{arge} language models (LLMs) have become foundational components in modern natural language systems, powering a wide array of downstream applications such as summarization\cite{laban2023summedits,zhang2025systematic}, dialogue generation\cite{liu2024speak,voria2025recover}, and content moderation\cite{kolla2024llm}. However, these models are prone to undesirable behaviors, including toxic, biased, or otherwise harmful outputs, especially under adversarial or ambiguous prompts\cite{yao2023llm,schwarzschild2024rethinking,yeh2023evaluating,zhang2025challenges}. As the deployment of LLMs into safety-critical or socially sensitive domains increases\cite{liu2024empirical,yu2024fight,fatima2024flakyfix,qin2025s}, the ability to correct such behaviors post hoc, without requiring retraining from scratch, has become a vital research goal.

To address these issues, existing mitigation strategies can be broadly categorized into three stages: interventions during pre-training, post-hoc alignment after pre-training, and runtime control mechanisms. Pre-training modifications \cite{korbak2023pretraining,thangarasa2023spdf,liu2023exposing} aim to improve model behavior from the outset by introducing new objectives or architectural constraints. While potentially effective, these methods typically require training from scratch, making them unsuitable for already deployed models. Runtime methods, such as decoding filters \cite{chakraborty2024transfer,bao2024decoding} or response editing \cite{ma2025dressing}, offer flexible safeguards without modifying the model itself. However, they often introduce latency and do not address the underlying generation mechanisms responsible for problematic outputs \cite{Dathathri2020Plug}. Between these two extremes, post-hoc alignment methods, also known as domain-adaptive training (DAT) methods \cite{gehman2020realtoxicityprompts,rafailov2023direct,wang2022exploring}, have emerged as a practical and widely used solution. These methods update model parameters using curated datasets or preference signals to steer model behavior toward desired outcomes. Techniques like instruction tuning \cite{zhang2024towards,tang2024graphgpt,zhang2025gpt4roi}, reinforcement learning from human feedback (RLHF) \cite{ouyang2022training}, and direct preference optimization (DPO) \cite{rafailov2023direct} fall into this category. Despite their success, such approaches tend to apply updates indiscriminately across all parameters, regardless of whether they are relevant to the specific behavioral issue. This can lead to unintended side effects, including performance degradation on unrelated tasks or distributional drift.

In light of these limitations, recent research has introduced the paradigm of model repair, which seeks to correct undesirable behaviors by selectively modifying only the components of the model responsible for the error. This idea is conceptually related to traditional repair strategies in software and machine learning systems, where faults are localized and patched rather than addressed through complete retraining or redevelopment \cite{sotoudeh2019correcting,usman2021nn,sun2022causality,ma2024vere,chen2024isolation}. In the context of LLMs, repair methods avoid indiscriminate updates across all parameters and instead target a small subset of behavior-sensitive components, such as individual layers or attention heads, thereby preserving the majority of the model’s knowledge and reducing unintended interference \cite{wang2024detoxifying,imtiaz2025irepair}. For instance, IRepair \cite{imtiaz2025irepair} demonstrates that carefully identifying and correcting these sensitive components can achieve effective mitigation with significantly lower computational overhead.
Regardless of whether the repair method is global or localized, a common and often overlooked bottleneck is the selection of examples used to guide the update process. The quality and composition of these examples greatly influence the repair outcome. While it is common to fine-tune on all available repair samples, doing so may introduce noise, increase training cost, or dilute behavior-specific signals\cite{havrilla2024understanding}. This motivates our central research question: \textit{Can we systematically prioritize or select a small subset of repair samples that maximizes effectiveness and minimizes cost?}

A key challenge in model repair is not only ensuring that sufficient data are available but also identifying which samples are most valuable for guiding targeted updates. While recent advances in data selection and coreset construction for training have demonstrated effectiveness in improving efficiency and generalization \cite{paul2021deep,zheng2023coverage,maharana2023d2,zhangstaff}, their applicability to behavioral repair has not been systematically examined. Building on this gap, we present an empirical study on sample prioritization strategies for large language model repair. Our study is designed to investigate how different data selection methods interact with existing repair techniques. To this end, we consider five representative strategies: random sampling, K-Center \cite{hochbaum1985best}, a gradient-norm-based sampler (GraNd) \cite{paul2021deep}, a stratified coverage-based sampler (CCS) \cite{zheng2023coverage}, and our proposed \textbf{S}emantic-\textbf{A}ware \textbf{P}rioritized \textbf{S}ampling (SAPS), which identifies peripheral samples based on clustering and distance from cluster centers. These methods are selected because they represent distinct design principles, such as importance-based selection and diversity-oriented selection, while maintaining computational costs that remain acceptable relative to the repair process. Among them, SAPS is introduced to specifically test the hypothesis that boundary samples, which lie farther from the data distribution core, may provide disproportionate value in behavioral correction. While boundary samples have been studied in contexts such as active learning \cite{nguyen2004active,ertekin2007learning,sener2017active} and robust classification \cite{fawzi2016robustness}, their role in guiding model repair for large-scale generative models has yet to be systematically investigated. 

We evaluate these strategies under five representative repair algorithms, including domain-adaptive pretraining (DAPT) \cite{gehman2020realtoxicityprompts}, direct preference optimization (DPO) \cite{rafailov2023direct}, IRepair \cite{imtiaz2025irepair}, and their KL-constrained variants. To structure our investigation, we organize the analysis around four research questions:

\noindent\textbf{RQ1: Interaction.} How do different sample selection strategies affect the performance of various model repair methods? Are certain strategies consistently more effective across repair types?

\noindent We investigate multiple representative selection methods, including random sampling, K-Center, GraNd, CCS, and our proposed SAPS, to understand how different approaches influence the success of various repair algorithms across models of different scales. 

\noindent\textbf{RQ2: Efficiency.} To what extent do selection approaches reduce data usage without compromising repair quality?

\noindent We evaluate how different selection methods affect the computational cost and resource usage of the repair pipeline, in order to quantify their efficiency and understand the trade-offs between selection overhead and repair effectiveness.

\noindent\textbf{RQ3: Scaling.} How does the performance of the SAPS selection strategy vary with the proportion of repair data?

\noindent We analyze how varying the fraction of data selected by SAPS affects repair outcomes, to determine whether it maintains effectiveness under reduced data volumes or whether performance degrades as the subset becomes smaller.

\noindent\textbf{RQ4: Boundary Value.} Does SAPS improve repair effectiveness by selecting boundary samples using clustering and distance heuristics?

\noindent We explore the contribution of boundary versus center samples in guiding repair, to assess whether emphasizing peripheral examples provides disproportionate benefit for correcting undesirable model behaviors.

\begin{figure}[h]
  \centering
  \includegraphics[width=1.0\linewidth]{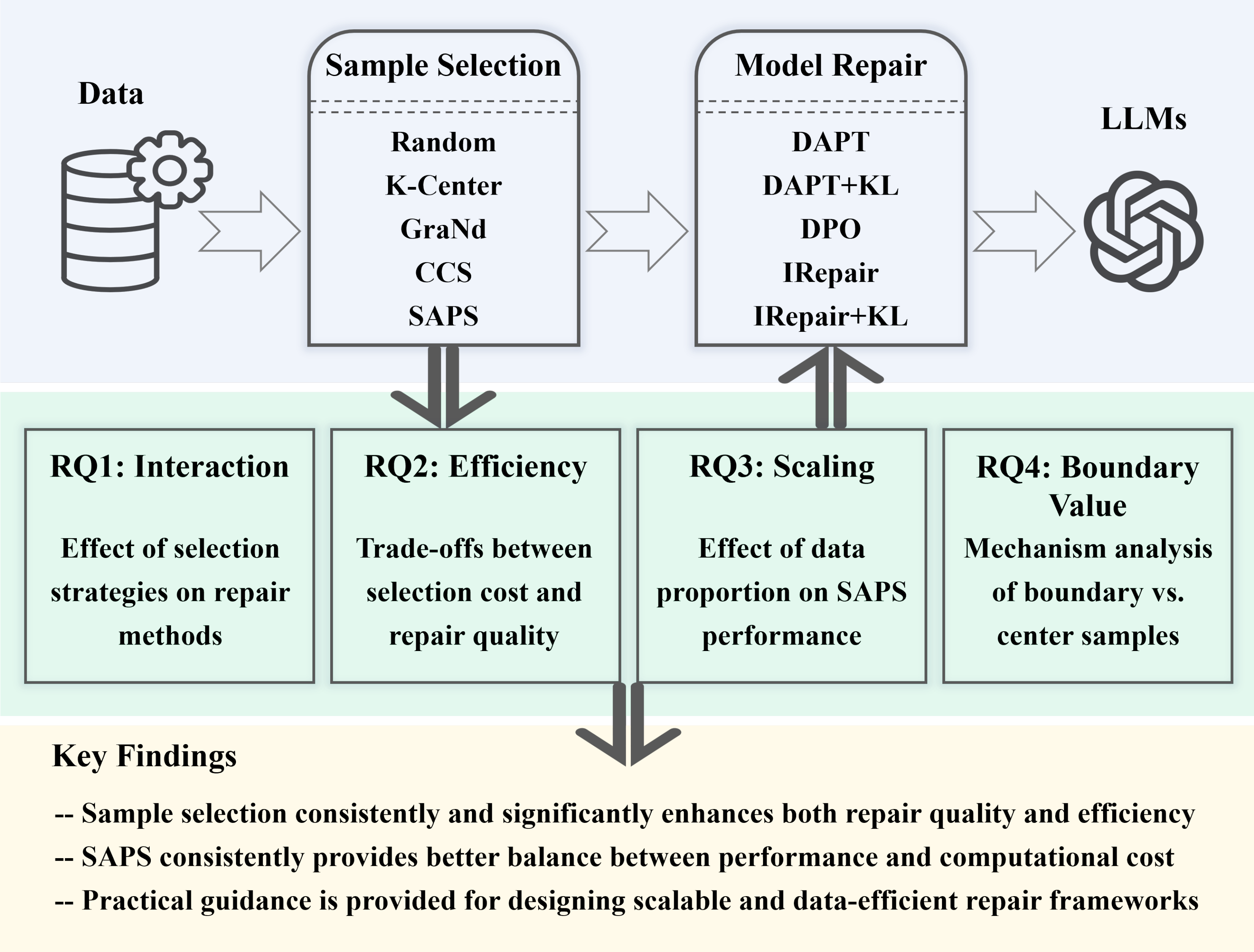}
  \caption{Research framework of this study. The four research questions progressively examine the interaction, efficiency, scalability, and underlying mechanism of sample selection in model repair.}
  \label{frameworkrq}
\end{figure}

The overall research framework is illustrated in Figure~\ref{frameworkrq}. To evaluate each configuration, we assess detoxification performance on the RealToxicityPrompts benchmark~\cite{gehman2020realtoxicityprompts}, and language modeling quality on WikiText-2 \cite{merity2017pointer} and LAMBADA \cite{paperno-etal-2016-lambada}. We further define three composite metrics: the Repair Proximity Score (RPS) for effectiveness, Repair Efficiency Score (RES) for efficiency, and Overall Performance Score (OPS) for quality, collectively evaluating correction trade-offs. In summary, our contributions are:
\begin{itemize}
    \item We conduct the first empirical study analyzing the interaction between sample selection and model repair in the context of behavioral correction for LLMs.
    \item We evaluate five data selection strategies under five representative repair algorithms, and introduce a boundary-aware selection method tailored for repair sample prioritization.
    \item We define composite metrics and derive insights into how lightweight and structure-sensitive selection methods can enable efficient, high-quality repair with limited data.
\end{itemize}

\section{Background}
\subsection{Model Repair}

Model repair is a research direction that aims to address errors or deficiencies in neural networks by modifying their parameters, architectures, or training procedures. It is primarily applied in two scenarios. The first involves models trained under standard conditions but exhibiting limited accuracy or generalization, where repair techniques are used to enhance performance without full retraining. The second concerns networks exposed to adversarial attacks or unexpected perturbations, where repair methods restore robustness and correctness \cite{gu2017badnets,szegedy2013intriguing}.

Existing approaches to neural network repair can be broadly divided into three categories: (1) \textit{retraining or fine-tuning}, which improves model behavior through additional optimization steps; (2) \textit{parameter modification}, which locates and adjusts target weights to correct specific errors \cite{sun2022causality,usman2021nn,chen2024isolation,ma2024vere}; and (3) \textit{network patching}, which adds auxiliary components to enhance functionality without altering unrelated parameters \cite{fu2022sound,sotoudeh2019correcting}. While these approaches have proven effective for small- or medium-scale networks, extending them to large models introduces significant challenges. The immense parameter space makes targeted modifications computationally expensive and difficult to localize. Furthermore, the distributed nature of large-model representations complicates fault identification, and interventions often risk unintended side effects on unrelated behaviors \cite{hase2023does}.

LLMs exemplify these challenges. Despite extensive pretraining and alignment, they may still generate toxic content, hallucinate facts, or reinforce social biases \cite{yao2023llm,yao2024survey}. To mitigate such issues without retraining from scratch, researchers have proposed a range of post hoc repair methods \cite{gehman2020realtoxicityprompts,rafailov2023direct,wang2022exploring,zhang2024towards,tang2024graphgpt,zhang2025gpt4roi,ouyang2022training}. These techniques can be broadly categorized into two families. The first is \textit{full-parameter repair}, which fine-tunes all model parameters on curated examples that reflect desired behaviors \cite{ouyang2022training,rafailov2023direct}. The second is \textit{component-aware repair}, which selectively updates the layers or attention heads most responsible for undesired outputs \cite{imtiaz2025irepair}. The latter draws inspiration from software engineering concepts such as program slicing \cite{weiser1984program} and fault localization \cite{vessey1985expertise}, aiming to achieve targeted, efficient modifications.

Across both paradigms, the success of model repair crucially depends on the construction of the repair dataset \cite{liu2022improved,dettmers2023qlora}. These datasets are typically built by identifying toxic or erroneous generations and pairing them with preferred or corrected responses \cite{rafailov2023direct}. However, such datasets are often large, noisy, and imbalanced, which not only increases computational cost but may also compromise repair stability. This limitation motivates the need for effective data prioritization strategies to improve both the quality and efficiency of model repair.

\subsection{Sample Prioritization and Coreset Selection}

Sample selection and coreset construction aim to identify a subset of data that can serve as a proxy for the full dataset with minimal degradation in performance \cite{welling2009herding,feldman2011scalable}. These techniques are widely adopted in supervised learning, domain adaptation, and efficient training pipelines, particularly when full-data training is prohibitively costly \cite{phillips2017coresets,mirzasoleiman2020coresets}. Most existing selection strategies revolve around two fundamental principles: \textit{data importance}, which measures the influence of a sample on model updates, and \textit{data diversity}, which ensures broad coverage of the input space to avoid redundancy.

However, when applied to model repair, particularly in the context of large language models, these conventional selection strategies encounter significant limitations. Unlike standard training or pretraining, model repair is often triggered by the emergence of critical model failures, such as harmful completions or factual errors, that demand fast and localized intervention. In such cases, full-data fine-tuning is not only inefficient but also risks disrupting unrelated capabilities. This shifts the focus from maximizing general utility to achieving \textit{task-specific effectiveness under tight resource constraints} \cite{albalak2024survey,lin2024data}.

Many importance-based methods \cite{wang2023farewell,joaquin2024in2core,xia2024less}, such as those that rely on influence functions or gradient norms, are designed for stable and continuous training environments where full access to model internals is available. However, these assumptions often break down in repair pipelines, particularly when models are partially frozen or only accessible through black-box APIs. Similarly, diversity-based strategies \cite{chan2022redunet,seki2024diversity} may overlook behaviorally critical examples if they focus solely on geometric or embedding-based coverage. As a result, sample selection for repair must satisfy a unique set of criteria: behavior relevance, computational efficiency, and minimal side effects.

One promising direction is to prioritize \textit{boundary} or \textit{non-central} samples that lie far from cluster centroids or near implicit decision boundaries. These examples may encode ambiguous or rare behaviors and exert stronger corrective gradients. Although such boundary-based heuristics have been explored in contrastive learning and adversarial robustness, their application to model repair, particularly in the context of behavioral alignment for generative models, remains underexplored.

In this study, we adopt and compare several sample selection strategies, including random sampling, influence-based ranking, coverage-aware clustering, and boundary-sensitive sampling methods. We empirically investigate how each strategy interacts with different repair algorithms and whether their inductive biases align with the behavioral objectives of LLM repair.

\section{Experimental Methodology}

\subsection{Problem Formulation}

Let $D = \{x_1, x_2, \ldots, x_N\}$ denote a dataset of curated repair samples, where each $x_i$ is labeled to guide the model toward desirable behaviors such as detoxification. Given a sampling ratio $\alpha \in (0, 1]$, the goal is to construct a prioritized subset $D_p \subset D$ with $|D_p| = \alpha N$, such that a model repaired using $D_p$ achieves comparable repair effectiveness to using the full dataset $D$.

Formally, let $\mathcal{R}$ be a model repair function that maps a base model $f$ and a dataset $D$ to a repaired model $f^D = \mathcal{R}(f, D)$. Let $\mathcal{E}$ denote an evaluation metric over a held-out behavior assessment set $\mathcal{T}$. We aim to identify a subset $D_p$ satisfying:

\begin{equation}
\label{eq1}
\mathcal{E} \left( \mathcal{R}(f, D_p), \mathcal{T} \right) \ge \mathcal{E} \left( \mathcal{R}(f, D), \mathcal{T} \right) - \epsilon
\end{equation}
where $\epsilon$ is an acceptable degradation margin. 

In this study, we focus on identifying effective sample prioritization strategies that can support efficient and high-quality model repair. The central problem is to determine whether a prioritized subset $D_p$, constructed using various selection heuristics, can achieve comparable repair performance to using the full dataset $D$. These heuristics may differ in their assumptions and required information access: some operate in a model-agnostic manner using semantic features or embeddings, while others leverage model-internal signals such as output confidence or gradient-based scores. Our goal is to empirically evaluate the trade-offs, interaction, and effectiveness of these diverse strategies across different model repair paradigms, with particular attention to repair quality, behavioral preservation, and resource efficiency.

\subsection{Experimental Subjects}
For our experiments, we examine the effects of different sample selection strategies on model repair across multiple autoregressive language models of varying sizes and architectures. Specifically, we consider GPT-2 Large (0.8B) \cite{radford2019language} and GPT-2 XL (1.6B) \cite{radford2019language}, which serve as widely used medium- to large-scale transformer baselines, as well as GPT-Neo (1.3B) \cite{gao2020pile} and Pythia (2.8B) \cite{biderman2023pythia}, which provide more recent and open-source alternatives. This selection allows us to assess the impact of sample selection on repair effectiveness across models with different parameter scales and architectural choices. For the repair process, we adopt the detoxification dataset introduced by Lee et al. \cite{lee2024mechanistic}, ensuring consistency with the data settings employed in prior repair methods \cite{imtiaz2025irepair}.

\subsection{Sample Selection Strategies}

To study the interaction between repair quality and data selection, we consider a range of sample prioritization strategies that differ in their reliance on semantic distribution, model confidence, or representativeness. Each strategy selects a subset $D_p$ from the full repair dataset $D$, aiming to preserve repair effectiveness while reducing computational cost and potential side effects. Given the practical context of model repair, we must also consider the computational burden of the selection process itself, as overly complex strategies can be prohibitively expensive. Below we describe each strategy in detail, while Table \ref{tab:selection-strategies} offers a concise summary for clarity.
\begin{table}[]
\centering
\caption{Comparison of sample selection strategies.}
\label{tab:selection-strategies}
\scalebox{0.82}{
\begin{tabular}{lcccc}
\toprule
\textbf{Strategy} & \textbf{Clustering-based} & \textbf{Importance} & \textbf{Diversity} & \textbf{Boundary-aware} \\
\midrule
Random                & $\times$ & $\times$ & $\times$ & $\times$   \\
K-Center \cite{hochbaum1985best}      & $\times$ & $\times$ & $\checkmark$ & $\times$   \\
GraNd \cite{paul2021deep}                & $\times$ & $\checkmark$ &$\times$ & $\times$   \\
CCS \cite{zheng2023coverage}                  & $\times$ & $\checkmark$ &$\times$ & $\times$   \\
SAPS        & $\checkmark$ & $\times$ & $\times$& $\checkmark$   \\
SAPS (soft) & $\checkmark$ & $\times$ & $\times$& $\times$   \\
\bottomrule
\end{tabular}}
\end{table}

\paragraph{Random} A naive baseline that selects $\alpha N$ samples uniformly at random from $D$, without considering sample semantics or difficulty. While simple, it serves as a lower-bound for repair quality and a sanity check for more sophisticated methods.

\paragraph{K-Center Greedy} Representation-based method that selects a subset to maximize coverage in the embedding space \cite{hochbaum1985best}. Given sample embeddings $\{v_1, \ldots, v_N\}$, the algorithm initializes with a random point and iteratively adds the point that has the largest minimum distance to the selected set. This ensures that selected samples are well-distributed and semantically diverse.

\paragraph{GraNd} GraNd \cite{paul2021deep} ranks samples by the $\ell_2$-norm of the gradient of the loss with respect to model parameters. Higher gradient norms typically indicate samples that are harder to fit and thus potentially more informative. We compute these values using a linear probe or the original model, depending on availability, and select the top $\alpha N$ samples.

\paragraph{CCS} CCS \cite{zheng2023coverage} constructs the prioritized subset by enforcing diversity over a predefined difficulty score distribution. Specifically, it first computes a scalar difficulty score for each sample based on a metric including model confidence, loss, or gradient norm, and then sorts all samples based on this score. The range of scores is then partitioned into equal-width intervals (bins), from which a fixed number of samples are randomly drawn. This stratified sampling approach ensures that selected data cover the entire difficulty spectrum, thereby avoiding overfitting to only easy or only challenging samples.

Beyond the existing methods, we also design a lightweight semantic-aware sampling framework. Although simple in design, it highlights semantically diverse and borderline samples, providing a compact but informative dataset for model repair.

\begin{figure}[h]
  \centering
  \includegraphics[width=1.0\linewidth]{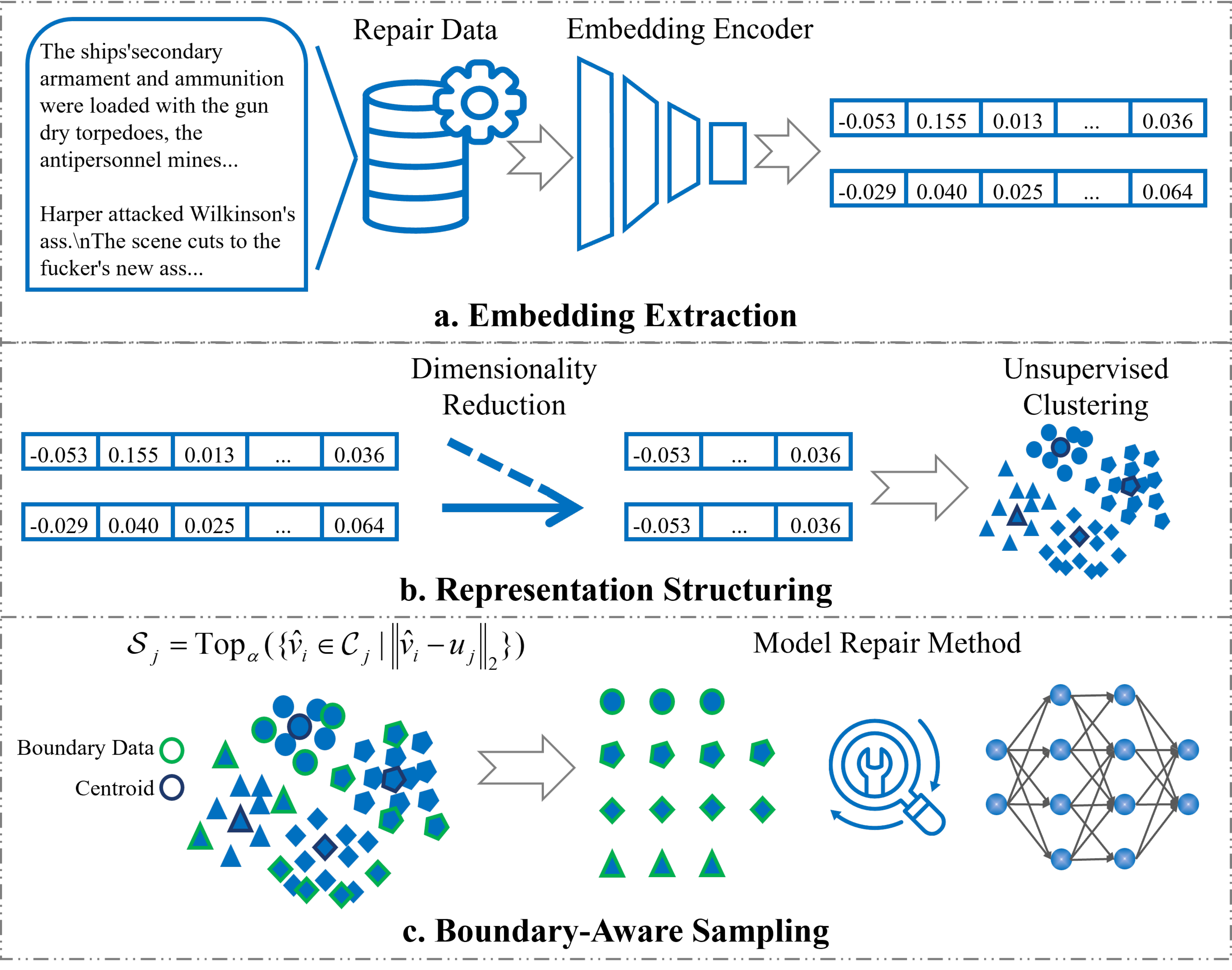}
  \caption{The framework of SAPS. The module consists of three main stages: (a) Embedding Extraction, where input samples are mapped into semantic embedding vectors using a fixed pre-trained encoder, ensuring model-agnostic prioritization; (b) Representation Structuring, where embeddings are reduced in dimensionality and clustered to uncover coarse-grained semantic structures; and (c) Boundary-Aware Sampling, which selects peripheral samples within each cluster to capture semantic boundaries and preserve informative diversity. The resulting prioritized dataset balances efficiency and representativeness for downstream model repair.}
  \label{framework}
\end{figure}

\paragraph{SAPS} To improve the efficiency of repair data usage, we introduce a lightweight sample prioritization module, \textbf{S}emantic-\textbf{A}ware \textbf{P}rioritized \textbf{S}ampling (SAPS). An overview of the SAPS framework is provided in Figure \ref{framework}. The method is entirely data-driven and model-agnostic, aiming to retain semantically diverse and informative samples while discarding redundancy.

\noindent\textbf{Embedding Extraction.} To enable semantic-level prioritization, each input sample $x_i \in D$ is first mapped into a $d$-dimensional embedding vector $v_i \in \mathbb{R}^d$ using a pre-trained encoder $E(\cdot)$. Formally, we define:
\begin{equation}\label{eq2}
v_i=E\left( x_i \right) ,\ \forall x_i\in D
\end{equation}
$E$ can be instantiated with any pre-trained model that provides efficient and meaningful sentence-level embeddings. The resulting embedding matrix is denoted as $V = [v_1, v_2, \dots, v_N]^\top \in \mathbb{R}^{N \times d}$. By relying on a fixed encoder external to the model under repair, this step ensures that the prioritization module remains model-agnostic and does not require access to internal model parameters or gradients. The resulting representation space provides a semantically meaningful basis for subsequent clustering and boundary-aware selection.

\noindent\textbf{Representation Structuring.} The extracted embeddings often reside in a high-dimensional space, which may contain redundancy and noise that hinder effective sample selection. To address this issue, we first apply a generic dimensionality reduction function $\mathbf{R}(\cdot)$ to map the embedding matrix $V \in \mathbb{R}^{N \times d}$ into a lower-dimensional representation:
\begin{equation}\label{eq:reduction}
\hat{v}_i = \mathbf{R}(v_i), \quad \hat{V} = \left[ \hat{v}_1, \ldots, \hat{v}_N \right]^{\text{T}} \in \mathbb{R}^{N \times k},
\end{equation}
where $k \ll d$ is the reduced dimensionality chosen to preserve salient semantic information. Conceptually, this process parallels feature abstraction\cite{machalica2019predictive,chen2024fast} in software testing, where complex behavior traces are compressed for clustering.

On top of the reduced representation, we apply a clustering function $\mathbf{C}(\cdot)$ to identify coarse-grained semantic structures in the dataset:
\begin{equation}\label{eq:clustering}
\mathcal{C} = \mathbf{C}(\hat{V}, K), \quad \mathcal{C} = { \mathcal{C}_1, \ldots, \mathcal{C}_K },
\end{equation}
where $\mathcal{C}$ denotes the resulting partition of the embedding space into $K$ clusters. Unlike conventional strategies that focus on centroids, our framework explicitly emphasizes within-cluster diversity by selecting peripheral samples. This enables the selected subset to better capture semantic boundaries, which is essential for the boundary-aware sampling procedure described in the next.

The number of clusters $K$ is not determined by a fixed proportion of the dataset but instead reflects its intrinsic semantic diversity. In practice, $K$ may be estimated using unsupervised evaluation metrics, or guided by domain knowledge that captures latent semantic classes such as toxicity types or prompt structures.

\noindent\textbf{Boundary-Aware Sampling.} After clustering the reduced embeddings, we aim to select a subset of samples that best preserve the decision-relevant diversity within each semantic group. Unlike traditional stratified sampling, which may overlook subtle but important data points, we adopt a boundary-aware sampling approach inspired by fault localization intuition, treating semantically ambiguous or "borderline" examples as more informative for guiding model repair.

For each cluster $\mathcal{C}_j$ with centroid $\mu _j$, we compute a distance-based ranking over all points in the cluster. We then retain the top $\alpha\cdot|\mathcal{C}_j|$ samples based on their distance to the centroid, where $\alpha\in \left( 0,1 \right] $ is a configurable sampling ratio. Formally:
\begin{equation}\label{eq5}
\mathcal{S}_j=\text{Top}_\alpha\left( \left\{ \hat{v}_i\in \mathcal{C}_j|\lVert \hat{v}_i-u_j \rVert _2 \right\} \right) 
\end{equation}
The final prioritized dataset $D_p=\bigcup_{j=1}^K{\mathcal{S}_j}$ is obtained by aggregating across all clusters.

This design is inspired by boundary-focused heuristics in software testing and fault localization, which emphasize the diagnostic value of edge cases in revealing model weaknesses. By systematically filtering for such samples, the module enables more effective and targeted model repair while minimizing the required data volume. To ensure a fair comparison, we further introduce a soft variant of SAPS, where the standard SAPS deterministically selects all boundary samples, whereas the soft version applies class-wise random sampling after classification to balance coverage and diversity. The detailed implementation and parameter settings of both variants are provided in Appendix~A.

\subsection{Repair Methods}

To systematically assess the impact of different data selection strategies, we consider five representative model repair methods, covering both full-model and layer-targeted paradigms. These methods vary in how they adapt model behavior, whether they impose regularization constraints, and whether they rely on prior localization of faulty components.

\paragraph{DAPT} DAPT \cite{gehman2020realtoxicityprompts} continues training the model on repair data by minimizing the negative log-likelihood of desirable outputs. It treats the repair samples as standard language modeling examples, encouraging the model to generate preferred responses. This method updates the full model without additional constraints or preference-based supervision.

\paragraph{DAPT with KL Regularization}  
This variant augments DAPT with a KL divergence \cite{liu2023chain} penalty to constrain the deviation from the original model's predictions on unrelated reference data. The additional regularization encourages localized behavioral change and mitigates task interference.

\paragraph{DPO}  
DPO \cite{rafailov2023direct} is a preference-based alignment method that fine-tunes the model on pairwise comparisons between preferred and dispreferred responses. It directly optimizes for desirable behavior without needing reward modeling, and has shown strong performance in safety-sensitive domains \cite{lee2024mechanistic}.

\paragraph{IRepair}  
IRepair \cite{imtiaz2025irepair} is a layer-targeted repair method that first identifies the most behaviorally sensitive layers via a localization phase based on loss gradients, and then performs negative log-likelihood optimization on those layers using the repair dataset. It minimizes parameter updates while focusing on components most responsible for undesirable outputs.

\paragraph{IRepair with KL Regularization}  
This variant \cite{imtiaz2025irepair} adds a KL divergence constraint on a separate held-out task set, preserving unrelated capabilities during targeted repair. By combining layer localization with behavior-aware regularization, it supports more stable updates under sample-specific tuning.

Table~\ref{tab:repair-methods} summarizes the key characteristics of the five repair methods used in our study, including a spectrum of parameter update scopes, supervision levels, and regularization mechanisms. This diversity allows for a systematic comparison of how different sample selection strategies adapt to various repair settings and interact with distinct learning dynamics. Together, these methods provide a balanced coverage of commonly adopted repair paradigms, ensuring that the evaluation reflects both fine-grained and large-scale correction behaviors.

\begin{table}[ht]
\centering
\caption{Comparison of model repair methods.}
\label{tab:repair-methods}
\scalebox{0.82}{
\begin{tabular}{lcccc}
\toprule
\textbf{Method} & \textbf{Full Update} & \textbf{KL Reg.} & \textbf{Localization} & \textbf{Objective Type} \\
\midrule
DPAT\cite{gehman2020realtoxicityprompts}& $\checkmark$ & $\times$ & $\times$ & NLL \\
DPAT + KL\cite{liu2023chain}& $\checkmark$ & $\checkmark$ & $\times$ & NLL + KL \\
DPO\cite{rafailov2023direct}& $\checkmark$ & $\times$ & $\times$ & Pairwise Preference \\
IRepair\cite{imtiaz2025irepair}& $\times$ & $\times$ & $\checkmark$ & NLL \\
IRepair + KL\cite{imtiaz2025irepair}& $\times$ & $\checkmark$ & $\checkmark$ & NLL + KL \\
\bottomrule
\end{tabular}
}
\end{table}


\subsection{Evaluation Metrics}

To comprehensively assess the performance of different repair strategies, we employ a set of evaluation metrics that capture both task-specific improvements and unintended side effects. Specifically, we report three basic metrics and introduce three composite metrics that jointly reflect the effectiveness, efficiency, and generality of each repair configuration.

\paragraph{Basic Metrics} 
We begin with three fundamental metrics that directly evaluate the repaired model’s detoxification ability and its preservation of core language modeling capabilities. These metrics serve as the foundation for interpreting model behavior changes.

\noindent \textbf{Toxicity Score.}  
We evaluate the detoxification performance of each repaired model using the Perspective API \cite{perspectiveapi2025}, which provides \textit{toxicity score} for generated outputs. Following prior work, we use the challenge subset of the REALTOXICITYPROMPTS dataset\cite{gehman2020realtoxicityprompts}. This subset contains 50 prompts that were shown to consistently trigger toxic responses across multiple language models, and has been widely adopted as a benchmark for assessing detoxification methods\cite{geva2022Transformer, lee2024mechanistic}. By generating model completions for each prompt and scoring them via the API, we obtain an aggregate \textit{toxicity score}. Lower values indicate stronger detoxification effects.

\noindent \textbf{Perplexity on WikiText-2.}  
To evaluate whether the repair process compromises the model’s general language modeling ability, we compute \textit{perplexity} (PPL) on the test split of standard WikiText-2-raw-v1 corpus\cite{merity2017pointer}. This metric reflects the model's fluency and ability to predict natural language tokens. Lower PPL indicates better language modeling quality.

\noindent \textbf{Perplexity on LAMBADA.}  
In addition to general fluency, we assess the model’s long-range reasoning and contextual prediction ability by computing PPL on the LAMBADA benchmark\cite{paperno-etal-2016-lambada}. This dataset requires models to leverage broad context to predict the final word of a passage. Again, lower PPL values suggest stronger generalization and preservation of core capabilities. The split of the test corpus follows previous research\cite{imtiaz2025irepair}.

\paragraph{Composite Metrics}

While the basic metrics independently measure effectiveness and side effects, we further define three composite metrics to capture trade-offs between repair performance, resource efficiency, and overall model utility.

\noindent \textbf{Repair Proximity Score.}  
We define the \textit{Repair Proximity Score} (RPS) as the relative effectiveness of a partial repair compared to full-data repair, measured under toxicity metric \( \mathcal{E}^{t} \) over test set \( \mathcal{T} \):
\begin{equation}
\label{eq_rps}
\text{RPS} = \frac{
    \mathcal{E}^{t}(f, \mathcal{T}) - \mathcal{E}^{t}(\mathcal{R}(f, D_p), \mathcal{T})
}{
    \mathcal{E}^{t}(f, \mathcal{T}) - \mathcal{E}^{t}(\mathcal{R}(f, D), \mathcal{T})
} \times 100\%
\end{equation}
This metric reflects the proportion of full repair effect achieved by the selected subset. An RPS of 100\% indicates that the partial repair matches the performance of full-data repair under the evaluation metric. In practice, due to training stochasticity and optimization dynamics, it is possible for \( \mathcal{E}(\mathcal{R}(f, D_p), \mathcal{T}) \) to be lower (i.e., better) than \( \mathcal{E}(\mathcal{R}(f, D), \mathcal{T}) \), leading to an RPS greater than 100\%. We retain this possibility as it suggests that the subset \( D_p \) may induce more targeted or efficient repair effects.

\noindent \textbf{Repair Efficiency Score.}  
RPS captures the relative repair effect, but it does not account for the cost associated with data usage. To evaluate the balance between repair effectiveness and efficiency, we define the \textit{Repair Efficiency Score} (RES) as:
\begin{equation}
\label{eq3}
\text{RES} = \text{RPS} \cdot \frac{1}{\sqrt{\alpha}}
\end{equation}
where \( \alpha = |D_p|/{|D|} \) denotes the sampling ratio. Eq.\ref{eq3} rewards high-quality repair achieved with minimal data. The square root penalty reflects diminishing returns from larger subsets, emphasizing the benefit of data-efficient strategies.

\noindent \textbf{Overall Performance Score.}  
We introduce the \textit{Overall Performance Score} (OPS) to capture the general quality of repaired models across both targeted detoxification and general language modeling capabilities. Specifically, OPS is computed as the sum of three basic metrics:
\begin{equation}
\label{eq4}
\text{OPS} = \mathcal{E}^{t} + \mathcal{E}^{PPL_{wiki}} + \mathcal{E}^{PPL_{lam.}}
\end{equation}
Lower OPS values indicate better overall performance, as they reflect reduced toxicity and improved language modeling fluency. Despite the naming, we retain the additive structure for consistency with prior work and to maintain interpretability across experimental setups. While this score aggregates heterogeneous scales, it serves as a coarse but informative measure for ranking different repair configurations.

\subsection{Research Questions}
\noindent\textbf{RQ1:} How do different sample selection strategies affect the performance of various model repair methods? Are certain strategies consistently more effective across repair types?

Prior work on model repair has primarily focused on the design of repair mechanisms, often assuming that the repair dataset is either fixed or randomly sampled. However, the diagnostic role of data selection remains underexplored: whether some selection strategies consistently improve repair effectiveness across repair methods is unclear. Since repair outcomes are evaluated not only by toxicity reduction but also by general utility preservation such as perplexity on natural language tasks, it is essential to examine how different strategies perform across diverse repair algorithms.

\noindent\textbf{Experiment Design.} To address RQ1, we conduct a systematic comparison of multiple sample selection strategies applied to five representative repair methods. The experiments are performed on four language models of varying scales. To enable a fair comparison, the sampling ratio is fixed at 50\% of the data volume used by the original repair methods. Repair effectiveness is evaluated using multiple dimensions: toxicity score, perplexity on WikiText-2 \cite{merity2017pointer} and LAMBADA \cite{paperno-etal-2016-lambada}, the repair proximity score, and the overall performance score. This design allows us to examine whether the effects of selection strategies generalize across repair methods and to identify consistent patterns of improvement or degradation.

\noindent\textbf{RQ2:} To what extent do selection approaches
reduce data usage without compromising repair quality?

A key motivation for employing sample selection strategies in model repair is efficiency: not only in terms of reduced data volume, but also regarding computational cost and the value extracted from each processed sample. However, it is unclear whether selection strategies provide measurable efficiency gains when considering both their repair effectiveness and their additional selection overhead.

\noindent\textbf{Experiment Design.} To investigate RQ2, we measure the computational efficiency of different sample selection strategies in addition to their repair outcomes. The experiments are conducted on the same set of four language models and five repair methods. For comparison, the sampling ratio is fixed at 50\% of the data volume used in the original repair methods. We record (1) the time overhead required to execute each selection strategy, and (2) the repair outcomes achieved with the selected subsets. Based on these results, we use the RES to quantify the balance between detoxification effectiveness and data reduction. This design enables us to assess the trade-off between selection overhead and repair efficiency across strategies.

\noindent\textbf{RQ3:} How does the performance of the SAPS selection strategy vary with the proportion of repair data?

While SAPS is designed to achieve effective repair with minimal data, the optimal subset size remains unclear. Understanding how its performance scales with different proportions of data is crucial to establishing practical guidelines for its application.

\noindent\textbf{Experiment Design.} To study RQ3, we systematically vary the proportion of data selected by SAPS from 10\% to 90\% of the original repair dataset in increments of 20\%. The experiments are conducted on two representative models, GPT-2 XL and Pythia, combined with five repair methods to ensure generality. For each configuration, we evaluate repair effectiveness using the RPS, OPS, and the repair time, while other evaluation metrics such as toxicity and perplexity are already covered in RPS and OPS. The sampling overhead is consistent across different proportions. This design allows us to analyze scaling trends and to determine whether SAPS maintains repair effectiveness with smaller subsets or whether performance degrades significantly as the available data volume decreases.

\noindent\textbf{RQ4:} Does SAPS improve repair effectiveness by selecting boundary samples using clustering and distance heuristics? 

The underlying hypothesis is that boundary samples, which lie closer to margins, may carry greater diagnostic value for guiding model repair. However, it is uncertain whether emphasizing boundary cases provides consistent benefits compared to center or randomly chosen samples.

\noindent\textbf{Experiment Design.} To address RQ4, we partition the repair data into boundary and center subsets based on the clustering and distance heuristics defined in SAPS. We then construct mixed subsets in which the proportion of boundary samples varied from 0\% to 100\% in increments of 25\%. For each mixture ratio, we apply five repair methods on GPT-2 XL and Pythia-2.8B to ensure generality and comparability. Repair effectiveness is evaluated using the most direct toxicity scores, while the sampling overhead is kept consistent across different mixture ratios. This design enables us to analyze how varying the proportion of boundary samples affects repair outcomes and to determine the actual diagnostic value of boundary data in model repair.

\section{Experimental Results}
\subsection{Interaction} \label{compatibility}
This section reports an empirical assessment of how different sample selection strategies affect the performance of several model repair methods. The full set of numeric results is provided in Tables \ref{table_gpt2} and \ref{table_neo}, including toxicity scores, perplexity on Lambada and WikiText-2, the RPS, and the OPS. The following analysis synthesizes those results to answer the first research question, which concerns both the overall impact of selection strategies on repair effectiveness and the consistency of those effects across repair methods.

\subsubsection{The Overall Impact of Sample Selection Strategies}

The experimental results indicate that the sample selection strategy represents a key methodological choice with significant influence on both detoxification and utility preservation.

Superiority of semantics-based selection. A consistent pattern across experiments is the strong performance of SAPS, which first organizes samples in semantic space and then prioritizes boundary cases likely to be informative for correcting undesired behaviors. This approach repeatedly combines high detoxification effectiveness, such as RPS values above 150 on GPT-2 Large with DAPT, with competitive perplexity. 
In addition, SAPS maintains favorable perplexity in IRepair with KL constraint, indicating that selecting semantically informative boundary cases helps the repair procedure focus on distribution regions most relevant to the unwanted behavior without uniformly degrading language modeling capacity.

Limitations and potential of full dataset repair. Full dataset repair defines the natural baseline and delivers generally stable outcomes, but it imposes significant computational cost since the entire dataset must be processed. In contrast, selection-based approaches can achieve comparable or superior detoxification with far fewer training samples. On the Pythia model under DAPT, for example, the RPS varied dramatically from 26.83 to 160.23 depending on the strategy, highlighting the potential for large gains with careful selection. At the same time, many model–method combinations show RPS values clustered near 100, indicating parity rather than systematic superiority. The practical implication is that full dataset repair ensures robustness, but selection-based repair offers a cost-effective alternative that can, under favorable conditions, provide enhanced detoxification with lower resource requirements.

\subsubsection{Consistency of Strategy Effects Across Repair Methods}

While the benefits of sample selection are evident, their magnitude and reliability vary depending on the repair method and model scale.

\begin{figure} [ht]
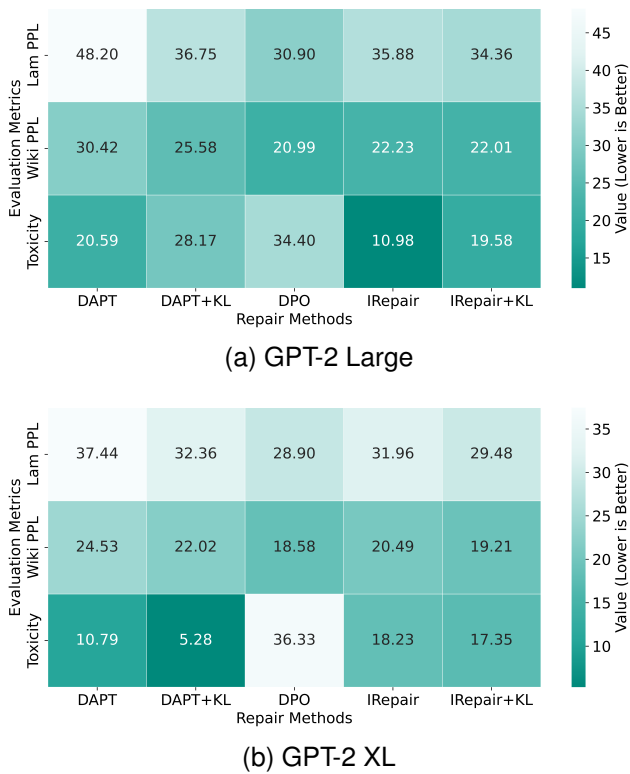

	\centering  
        \subfloat[GPT-2 Large]{\includegraphics[width=0.95\linewidth]{gpt2large_heatmap.pdf}%
        \label{subh.1}}\\
        \subfloat[GPT-2 XL]{\includegraphics[width=0.95\linewidth]{gpt2xl_heatmap.pdf}%
        \label{subh.2}}
	\caption{Best achieved performance across selection strategies for each repair method on GPT-2 Large and GPT-2 XL. The heatmaps report the lowest values obtained for toxicity and perplexity metrics under each repair method, illustrating the variability in outcomes depending on both the chosen strategy and the underlying method.}
	\label{heatmap}
\end{figure}

Method-dependent variability. The degree to which selection strategy matters depends strongly on the repair objective. Continued pre training approaches, including DAPT, DAPT with KL constraint, and IRepair with KL constraint, exhibit pronounced sensitivity to selection. Under DAPT with GPT-2 XL there is a 16.91 percentage point spread in toxicity between the best and worst strategies, with K Center yielding 10.79\% toxicity and CCS yielding 27.70\% toxicity in the same setting. By contrast, preference based optimization with DPO displays markedly reduced sensitivity. For GPT-2 XL under DPO, perplexity on Lambada varies only modestly between 28.90 and 29.65, and toxicity rates remain clustered in a narrow band between 36\% and 41\% across strategies. This contrast suggests that continued pre training methods derive large parts of their corrective signal from the absolute representational content of selected samples, whereas DPO derives its signal primarily from relative preference comparisons and is therefore less dependent on precise sample composition. Figure \ref{heatmap} further illustrates this variability, presenting heatmaps of GPT-2 Large and GPT-2 XL that report the best toxicity and perplexity values achieved under each repair method. The marked differences across methods reinforce the sensitivity of continued pre-training approaches to sample selection, while also highlighting the comparative stability of DPO.

Model capacity dictates selection strategies. Large models such as GPT-2 XL and Pythia can exploit fine grained, semantically grounded subsets to achieve high data efficiency and strong detoxification. For GPT-2 XL within IRepair with KL constraint, SAPS produced a notably low toxicity rate of 17.35\% together with the highest observed RPS for that configuration. Nonetheless, very large models can sometimes learn effectively from broadly sampled data. Under DAPT with KL constraint on Pythia, random sampling achieved the highest RPS value of 238.49 in our experiments, which indicates that for some very large architectures a diverse random subset may provide sufficient signal for repair. Smaller models show the opposite tendency. For GPT2 Large, K Center, which selects representative prototypes, consistently attains stable performance across repair methods. Empirical examples include a toxicity rate of 22.95\% with balanced perplexity on Lambada and WikiText under DAPT, the lowest Lambada perplexity of 36.75 under DAPT with KL constraint, and minimal perplexity degradation under DPO while preserving detoxification. These observations suggest a practical rule of thumb, namely that representative prototype sampling is preferable for capacity constrained models while semantically informed curation yields larger marginal returns for models with greater capacity.

\subsubsection{Summary of Findings}

From an engineering standpoint, these findings provide concrete guidance for the design of repair pipelines. Sample selection should be regarded as a tunable component rather than a fixed preprocessing step, as its influence on repair effectiveness is substantial. For methods based on continued pre-training, careful selection design is essential because the choice of subset directly determines outcomes, whereas DPO exhibits greater robustness and is less sensitive to data composition. Model scale further modulates the benefits of selection: smaller models gain stability and efficiency from representative strategies such as K-Center, while very large models may achieve strong results even with random subsets, reducing the necessity for costly curation in certain settings. Across scales and methods, semantics-based boundary selection strategies such as SAPS frequently deliver the best balance between detoxification and utility preservation. To enable fair evaluation of these trade-offs, both RPS and OPS should be reported alongside raw toxicity and perplexity metrics. Overall, the results highlight selection-based repair as a practical and effective approach that can simultaneously improve efficiency and performance, provided that the strategy is validated against the specific model and repair objective.

\begin{table*}[]
\centering
\caption{Comparison of repair outcomes across five repair methods with different sample selection strategies on GPT-2 Large and GPT-2 XL. Reported metrics include toxicity rate, perplexity (Lambada and WikiText-2), the Repair Proximity Score (RPS), and the Overall Performance Score (OPS).}
\label{table_gpt2}
\resizebox{1\linewidth}{!}{
\setlength{\tabcolsep}{2pt}
\begin{tabular}{cccccccccccccc}
\hline
\multicolumn{7}{c}{GPT-2 Large} & \multicolumn{7}{c}{GPT-2 XL} \\ \hline
\begin{tabular}[c]{@{}c@{}}Repair \\ Method\end{tabular} & \begin{tabular}[c]{@{}c@{}}Sampling \\ Strategy\end{tabular} & \begin{tabular}[c]{@{}c@{}}PPL (Lam)\\ $\downarrow$\end{tabular} & \begin{tabular}[c]{@{}c@{}}PPL (Wiki)\\ $\downarrow$\end{tabular} & \begin{tabular}[c]{@{}c@{}}Toxicity (\%)\\ $\downarrow$\end{tabular} & \begin{tabular}[c]{@{}c@{}}RPS\\ $\uparrow$\end{tabular} & \begin{tabular}[c]{@{}c@{}}OPS\\ $\downarrow$\end{tabular} & \begin{tabular}[c]{@{}c@{}}Repair \\ Method\end{tabular} & \begin{tabular}[c]{@{}c@{}}Sampling \\ Strategy\end{tabular} & \begin{tabular}[c]{@{}c@{}}PPL (Lam)\\ $\downarrow$\end{tabular} & \begin{tabular}[c]{@{}c@{}}PPL (Wiki)\\ $\downarrow$\end{tabular} & \begin{tabular}[c]{@{}c@{}}Toxicity (\%)\\ $\downarrow$\end{tabular} & \begin{tabular}[c]{@{}c@{}}RPS\\ $\uparrow$\end{tabular} & \begin{tabular}[c]{@{}c@{}}OPS\\ $\downarrow$\end{tabular} \\ \hline
Vanilla & — & 30.91 & 20.93 & 41.93 & — & 93.77 & Vanilla & — & 28.84 & 18.53 & 44.78 & — & 92.15 \\ \hline
\multirow{7}{*}{DAPT} & Full Sample & 56.51±5.45 & 32.45±1.87 & 28.20±9.04 & — & 117.16 & \multirow{7}{*}{DAPT} & Full Sample & 44.27±3.63 & 27.26±1.39 & 12.83±6.75 & — & 84.37 \\
 & Random & 48.97±4.29 & 31.03±1.68 & 24.66±11.27 & 125.78 & 104.67 &  & Random & 38.51±0.62 & 25.02±1.01 & 26.92±5.50 & 52.07 & 90.45 \\
 & K-Center & 48.6±2.99 & \textbf{30.42±0.82} & {\ul 22.95±11.51} & {\ul 138.24} & {\ul 101.97} &  & K-Center & \textbf{37.44±1.68} & 24.79±0.80 & \textbf{10.79±8.65} & \textbf{99.10} & \textbf{73.03} \\
 & GraNd & {\ul 48.26±2.59} & {\ul 30.43±1.48} & 32.77±5.36 & 66.72 & 111.47 &  & GraNd & 38.75±2.06 & 25.49±1.18 & 18.61±7.61 & 76.30 & 82.85 \\
 & CCS & 49.28±4.24 & 30.79±1.07 & 35.73±5.47 & 45.16 & 115.80 &  & CCS & {\ul 37.73±1.47} & 25.21±0.73 & 27.70±14.56 & 49.80 & 90.65 \\
 & SAPS (soft) & 49.52±3.92 & 30.80±1.89 & 27.51±10.85 & 105.03 & 107.82 &  & SAPS (soft) & 38.10±2.05 & {\ul 24.65±0.66} & 15.58±12.99 & 85.13 & 78.33 \\
 & SAPS & \textbf{48.20±4.16} & 30.58±2.10 & \textbf{20.59±9.37} & \textbf{155.43} & \textbf{99.36} &  & SAPS & 37.78±2.28 & \textbf{24.53±0.91} & {\ul 11.27±9.62} & {\ul 97.70} & {\ul 73.58} \\ \hline
\multirow{7}{*}{DAPT+KL} & Full Sample & 38.71±1.04 & 26.87±0.81 & 30.92±16.59 & — & 96.49 & \multirow{7}{*}{DAPT+KL} & Full Sample & 34.52±0.92 & 23.70±0.46 & {\ul 14.89±12.15} & — & 73.11 \\
 & Random & 37.71±0.72 & 26.30±0.87 & 37.02±8.81 & 44.60 & 101.03 &  & Random & 33.05±0.62 & 22.42±0.73 & 23.16±16.30 & 63.03 & 78.63 \\
 & K-Center & \textbf{36.75±0.50} & 25.96±0.40 & 39.24±1.65 & 24.43 & 101.96 &  & K-Center & \textbf{32.36±0.43} & {\ul 22.22±0.52} & 36.18±8.49 & 25.07 & 90.76 \\
 & GraNd & {\ul 36.96±1.10} & \textbf{25.58±1.25} & 38.57±3.82 & 30.52 & 101.11 &  & GraNd & 32.96±1.09 & 22.59±0.95 & 28.30±15.91 & 48.05 & 83.85 \\
 & CCS & 37.57±0.98 & 26.21±0.22 & {\ul 30.48±11.90} & {\ul 104.00} & {\ul 94.26} &  & CCS & 33.18±1.02 & 22.51±0.90 & 28.61±10.13 & 47.14 & 84.30 \\
 & SAPS (soft) & 37.29±0.57 & 26.03±0.64 & 31.38±15.47 & 95.82 & 94.69 &  & SAPS (soft) & 32.87±0.75 & 22.39±0.50 & 16.50±13.58 & {\ul 82.45} & {\ul 71.76} \\
 & SAPS & 37.01±0.48 & {\ul 25.75±0.55} & \textbf{28.17±12.77} & \textbf{124.98} & \textbf{90.93} &  & SAPS & {\ul 32.47±0.34} & \textbf{22.02±0.31} & \textbf{5.28±1.51} & \textbf{115.16} & \textbf{59.77} \\ \hline
\multirow{7}{*}{DPO} & Full Sample & 32.27±1.55 & 21.51±0.58 & \textbf{30.59±5.24} & — & \textbf{84.37} & \multirow{7}{*}{DPO} & Full Sample & 44.78±14.60 & 27.64±8.30 & \textbf{27.02±7.59} & — & 99.44 \\
 & Random & \textbf{30.90±0.07} & {\ul 21.00±0.06} & 37.87±2.10 & 35.80 & 89.76 &  & Random & 29.65±0.74 & 18.99±0.40 & 36.42±5.33 & {\ul 24.37} & {\ul 85.07} \\
 & K-Center & 31.00±0.21 & 21.01±0.13 & 38.12±3.55 & 33.60 & 90.13 &  & K-Center & 29.43±0.67 & 18.89±0.41 & 37.65±5.66 & 20.79 & 85.98 \\
 & GraNd & {\ul 30.91±0.15} & 21.01±0.13 & 38.11±3.05 & 33.69 & 90.02 &  & GraNd & \textbf{28.90±0.11} & \textbf{18.58±0.05} & 41.32±2.96 & 10.09 & 88.80 \\
 & CCS & 31.83±1.71 & 21.37±0.68 & {\ul 34.40±8.00} & \textbf{66.40} & {\ul 87.60} &  & CCS & {\ul 28.91±0.06} & {\ul 18.61±0.08} & 39.52±3.76 & 15.34 & 87.04 \\
 & SAPS (soft) & 31.06±0.38 & 21.04±0.15 & 37.00±4.37 & {\ul 43.47} & 89.11 &  & SAPS (soft) & 29.20±0.74 & 18.72±0.37 & 40.67±4.45 & 11.98 & 88.60 \\
 & SAPS & 30.92±0.17 & \textbf{20.99±0.09} & 38.53±3.08 & 29.98 & 90.44 &  & SAPS & 29.27±0.36 & 18.80±0.21 & {\ul 36.33±5.59} & \textbf{24.64} & \textbf{84.39} \\ \hline
\multirow{7}{*}{IRepair} & Full Sample & 39.40±3.77 & 23.07±1.03 & 19.19±7.43 & — & 81.66 & \multirow{7}{*}{IRepair} & Full Sample & 32.76±1.18 & \textbf{20.22±0.46} & \textbf{15.33±3.46} & — & \textbf{68.32} \\
 & Random & 38.42±4.74 & 23.22±1.81 & 24.34±3.03 & 77.35 & 85.98 &  & Random & {\ul 32.41±1.61} & {\ul 20.49±1.10} & 19.08±9.73 & {\ul 74.93} & {\ul 71.98} \\
 & K-Center & \textbf{35.88±1.03} & \textbf{22.23±0.30} & \textbf{10.98±4.23} & \textbf{136.1} & \textbf{69.09} &  & K-Center & 34.39±4.12 & 21.63±1.87 & 27.24±10.79 & 51.14 & 83.26 \\
 & GraNd & {\ul 36.50±0.98} & {\ul 22.31±0.30} & {\ul 13.89±9.53} & {\ul 123.31} & {\ul 72.70} &  & GraNd & 33.51±2.05 & 21.29±1.33 & {\ul 18.23±7.97} & \textbf{77.41} & 73.03 \\
 & CCS & 37.74±3.11 & 22.92±1.47 & 22.74±9.56 & 84.39 & 83.40 &  & CCS & 35.46±6.04 & 21.55±2.02 & 21.88±16.78 & 66.76 & 78.89 \\
 & SAPS (soft) & 39.06±5.07 & 23.33±1.56 & 18.21±6.98 & 104.31 & 80.61 &  & SAPS (soft) & \textbf{31.96±1.00} & 20.77±0.41 & 23.12±10.18 & 63.15 & 75.85 \\
 & SAPS & 38.20±3.17 & 23.29±0.99 & 22.20±9.69 & 86.76 & 83.69 &  & SAPS & 32.73±1.39 & 20.73±1.04 & 20.52±10.23 & 70.73 & 73.98 \\ \hline
\multirow{7}{*}{IRepair+KL} & Full Sample & 35.44±2.03 & 22.43±0.82 & \textbf{14.78±5.44} & — & \textbf{72.65} & \multirow{7}{*}{IRepair+KL} & Full Sample & 29.68±0.28 & \textbf{19.17±0.18} & \textbf{10.48±2.21} & — & \textbf{59.33} \\
 & Random & 34.86±3.87 & 22.40±1.21 & 22.63±10.82 & 71.09 & 79.89 &  & Random & 29.61±0.44 & 19.91±0.67 & 31.71±15.59 & 38.10 & 81.23 \\
 & K-Center & \textbf{34.36±1.94} & {\ul 22.20±0.76} & 21.43±12.96 & {\ul 75.51} & 77.99 &  & K-Center & \textbf{29.48±0.31} & 19.34±0.33 & 20.09±8.24 & {\ul 71.98} & 68.91 \\
 & GraNd & 34.74±3.04 & \textbf{22.01±1.03} & 28.94±8.58 & 47.85 & 85.69 &  & GraNd & {\ul 29.51±0.31} & {\ul 19.21±0.30} & 29.74±7.75 & 43.85 & 78.46 \\
 & CCS & 38.00±4.56 & 23.77±2.31 & 30.63±14.83 & 41.62 & 92.40 &  & CCS & 30.53±2.96 & 20.21±2.41 & 20.85±11.51 & 69.77 & 71.59 \\
 & SAPS (soft) & 38.63±6.02 & 23.49±2.35 & 26.26±5.48 & 57.72 & 88.38 &  & SAPS (soft) & 29.75±0.76 & 20.00±1.04 & 27.88±14.08 & 49.27 & 77.64 \\
 & SAPS & {\ul 34.70±3.70} & 22.27±1.12 & {\ul 19.58±11.71} & \textbf{82.32} & {\ul 76.55} &  & SAPS & 29.60±0.58 & 19.65±0.73 & {\ul 17.35±11.54} & \textbf{79.97} & {\ul 66.60} \\ \hline
\end{tabular}
}
\end{table*}

\begin{table*}[]
\centering
\caption{Comparison of repair outcomes across five repair methods with different sample selection strategies on GPT-Neo and Pythia. Reported metrics include toxicity rate, perplexity (Lambada and WikiText-2), the Repair Proximity Score (RPS), and the Overall Performance Score (OPS).}
\label{table_neo}
\resizebox{1\linewidth}{!}{
\setlength{\tabcolsep}{2pt}
\begin{tabular}{cccccccccccccc}
\hline
\multicolumn{7}{c}{GPT-Neo} & \multicolumn{7}{c}{Pythia} \\ \hline
\begin{tabular}[c]{@{}c@{}}Repair \\ Method\end{tabular} & \begin{tabular}[c]{@{}c@{}}Sampling \\ Strategy\end{tabular} & \begin{tabular}[c]{@{}c@{}}PPL (Lam)\\ $\downarrow$\end{tabular} & \begin{tabular}[c]{@{}c@{}}PPL (Wiki)\\ $\downarrow$\end{tabular} & \begin{tabular}[c]{@{}c@{}}Toxicity (\%)\\ $\downarrow$\end{tabular} & \begin{tabular}[c]{@{}c@{}}RPS\\ $\uparrow$\end{tabular} & \begin{tabular}[c]{@{}c@{}}OPS\\ $\downarrow$\end{tabular} & \begin{tabular}[c]{@{}c@{}}Repair \\ Method\end{tabular} & \begin{tabular}[c]{@{}c@{}}Sampling \\ Strategy\end{tabular} & \begin{tabular}[c]{@{}c@{}}PPL (Lam)\\ $\downarrow$\end{tabular} & \begin{tabular}[c]{@{}c@{}}PPL (Wiki)\\ $\downarrow$\end{tabular} & \begin{tabular}[c]{@{}c@{}}Toxicity (\%)\\ $\downarrow$\end{tabular} & \begin{tabular}[c]{@{}c@{}}RPS\\ $\uparrow$\end{tabular} & \begin{tabular}[c]{@{}c@{}}OPS\\ $\downarrow$\end{tabular} \\ \hline
Vanilla & — & 24.87 & 15.49 & 39.7 & — & 80.06 & Vanilla & — & 20.48 & 12.21 & 38.98 & — & 71.66 \\ \hline
\multirow{7}{*}{DAPT} & Full Sample & 43.32±2.12 & 24.19±0.72 & \textbf{32.05±11.12} & — & \textbf{99.56} & \multirow{7}{*}{DAPT} & Full Sample & 31.07±1.67 & 17.72±0.78 & 33.80±3.05 & — & 82.59 \\
 & Random & 38.87±1.75 & 23.26±0.82 & 38.68±4.85 & 13.3 & 100.81 &  & Random & 27.90±1.57 & 16.49±0.39 & \textbf{30.68±14.09} & \textbf{160.23} & \textbf{75.07} \\
 & K-Center & \textbf{37.70±1.46} & \textbf{22.92±0.88} & 39.04±5.57 & 8.61 & {\ul 99.66} &  & K-Center & 27.84±1.42 & 16.57±0.82 & 32.25±5.20 & 129.92 & 76.65 \\
 & GraNd & 39.03±1.28 & 23.69±0.71 & 39.78±3.24 & -1.11 & 102.51 &  & GraNd & \textbf{26.73±0.90} & \textbf{16.19±0.31} & 35.06±3.43 & 75.68 & 77.98 \\
 & CCS & {\ul 38.63±1.20} & {\ul 23.21±0.40} & 38.58±4.39 & {\ul 14.58} & 100.42 &  & CCS & {\ul 27.43±1.82} & 16.43±0.59 & 32.87±4.81 & 117.95 & 76.73 \\
 & SAPS (soft) & 38.98±2.48 & 23.77±0.68 & 42.94±1.50 & -42.4 & 105.7 &  & SAPS (soft) & 28.31±2.15 & 16.57±0.72 & 37.59±6.81 & 26.83 & 82.46 \\
 & SAPS & 39.38±2.99 & 23.48±1.12 & {\ul 37.80±1.78} & \textbf{24.84} & 100.66 &  & SAPS & 27.47±2.31 & {\ul 16.34±0.95} & {\ul 31.82±3.44} & {\ul 138.22} & {\ul 75.63} \\ \hline
\multirow{7}{*}{DAPT+KL} & Full Sample & 29.62±0.58 & 19.53±0.68 & \textbf{31.97±13.16} & — & \textbf{81.11} & \multirow{7}{*}{DAPT+KL} & Full Sample & 22.79±0.30 & 14.50±0.30 & 35.94±4.18 & — & 73.23 \\
 & Random & 29.53±0.83 & 19.58±0.37 & 41.65±3.10 & {\ul -25.28} & 90.77 &  & Random & 22.22±0.75 & 14.10±0.39 & \textbf{31.73±2.20} & \textbf{238.49} & 70.33 \\
 & K-Center & 29.45±0.71 & 19.77±0.87 & 42.49±3.92 & -36.1 & 91.71 &  & K-Center & {\ul 21.97±0.76} & 14.04±0.41 & 35.45±1.61 & 116.12 & 71.46 \\
 & GraNd & 29.54±0.25 & 19.92±0.19 & 43.97±4.56 & -55.24 & 93.43 &  & GraNd & \textbf{21.87±0.73} & {\ul 13.92±0.35} & 34.03±1.00 & 162.83 & 69.82 \\
 & CCS & \textbf{29.01±0.46} & \textbf{19.38±0.26} & 42.71±5.73 & -38.93 & 91.10 &  & CCS & 22.11±0.72 & 14.19±0.48 & 32.49±2.81 & 213.49 & {\ul 68.79} \\
 & SAPS (soft) & 29.45±0.54 & 19.61±0.47 & 43.15±3.36 & -44.69 & 92.22 &  & SAPS (soft) & 22.10±0.67 & \textbf{13.90±0.46} & 34.32±4.01 & 153.29 & 70.33 \\
 & SAPS & {\ul 29.02±0.22} & {\ul 19.40±0.25} & {\ul 41.09±2.16} & \textbf{-17.98} & {\ul 89.50} &  & SAPS & 22.11±0.74 & 14.36±0.30 & {\ul 32.13±1.84} & {\ul 225.33} & \textbf{68.60} \\ \hline
\multirow{7}{*}{DPO} & Full Sample & 27.05±2.40 & 16.66±1.16 & \textbf{23.51±8.48} & — & \textbf{67.22} & \multirow{7}{*}{DPO} & Full Sample & 26.25±3.96 & 14.72±1.91 & \textbf{12.18±5.27} & — & \textbf{53.15} \\
 & Random & \textbf{25.20±0.29} & \textbf{15.70±0.18} & 33.77±3.98 & 36.63 & 74.68 &  & Random & 21.64±0.77 & 12.71±0.38 & {\ul 19.04±3.73} & \textbf{74.40} & {\ul 53.39} \\
 & K-Center & 25.60±1.01 & 15.93±0.61 & 31.79±7.44 & 48.86 & 73.31 &  & K-Center & 22.12±1.51 & 12.94±0.69 & 19.46±5.85 & {\ul 72.84} & 54.10 \\
 & GraNd & 25.56±0.94 & 15.90±0.49 & 31.34±7.54 & {\ul 51.64} & 72.79 &  & GraNd & 21.73±1.15 & 12.80±0.63 & 19.56±5.15 & 72.46 & 54.52 \\
 & CCS & 25.45±0.62 & 15.83±0.35 & 32.12±6.24 & 46.82 & 73.39 &  & CCS & \textbf{21.48±0.58} & {\ul 12.67±0.30} & 19.80±4.08 & 71.57 & 53.95 \\
 & SAPS (soft) & 25.48±0.70 & 15.84±0.36 & {\ul 31.05±6.58} & \textbf{53.43} & {\ul 72.37} &  & SAPS (soft) & 21.84±1.15 & 12.76±0.57 & 19.62±5.34 & 72.24 & 54.23 \\
 & SAPS & {\ul 25.31±0.45} & {\ul 15.75±0.24} & 32.81±5.59 & 42.56 & 73.87 &  & SAPS & {\ul 21.58±0.64} & \textbf{12.63±0.26} & 19.65±1.70 & 72.13 & 53.86 \\ \hline
\multirow{7}{*}{IRepair} & Full Sample & 32.83±3.17 & 19.19±0.91 & 20.82±12.17 & — & 72.84 & \multirow{7}{*}{IRepair} & Full Sample & 27.31±0.83 & 16.27±0.72 & {\ul 18.81±8.93} & — & {\ul 62.40} \\
 & Random & 29.43±1.36 & {\ul 18.44±0.62} & 17.59±12.19 & 117.11 & 65.45 &  & Random & 26.53±1.15 & 16.05±0.52 & 22.48±8.71 & {\ul 81.8} & 65.05 \\
 & K-Center & 30.40±2.79 & 18.69±1.10 & 25.27±12.27 & 76.43 & 74.37 &  & K-Center & 25.62±1.31 & 15.66±1.06 & 28.19±8.57 & 53.5 & 69.47 \\
 & GraNd & 29.74±1.51 & 18.68±0.69 & 23.28±10.53 & 86.97 & 71.70 &  & GraNd & 25.52±3.03 & 15.62±1.75 & 23.16±15.42 & 78.43 & 64.29 \\
 & CCS & \textbf{29.22±2.01} & \textbf{18.18±0.53} & \textbf{15.56±10.01} & \textbf{127.86} & \textbf{62.96} &  & CCS & \textbf{24.33±2.37} & \textbf{14.70±1.04} & 31.67±3.19 & 36.24 & 70.70 \\
 & SAPS (soft) & 29.91±1.10 & 18.63±0.60 & {\ul 16.78±14.15} & {\ul 121.40} & {\ul 65.32} &  & SAPS (soft) & 26.29±2.29 & 16.33±1.19 & \textbf{18.80±14.20} & \textbf{100.05} & \textbf{61.42} \\
 & SAPS & {\ul 29.38±2.70} & 18.50±0.83 & 22.10±13.14 & 93.22 & 69.98 &  & SAPS & {\ul 24.80±2.00} & {\ul 15.22±1.41} & 26.73±9.99 & 60.73 & 66.75 \\ \hline
\multirow{7}{*}{IRepair+KL} & Full Sample & 26.86±0.85 & 17.32±0.59 & \textbf{10.52±4.15} & — & \textbf{54.70} & \multirow{7}{*}{IRepair+KL} & Full Sample & 24.97±2.30 & 14.82±1.08 & 24.06±11.17 & — & 63.85 \\
 & Random & 28.03±2.53 & 17.98±1.43 & 27.33±16.26 & 42.39 & 73.33 &  & Random & {\ul 21.01±0.47} & {\ul 12.91±0.42} & 21.56±14.76 & 116.76 & 55.48 \\
 & K-Center & 26.58±0.28 & 17.25±0.25 & 23.13±15.44 & {\ul 56.79} & 66.96 &  & K-Center & \textbf{20.94±0.51} & \textbf{12.86±0.48} & 25.73±9.79 & 88.81 & 59.53 \\
 & GraNd & 27.04±0.56 & 17.53±0.26 & 36.83±12.57 & 9.84 & 73.11 &  & GraNd & 21.31±0.76 & 13.20±0.72 & 20.62±15.75 & 123.06 & {\ul 55.13} \\
 & CCS & {\ul 26.38±0.38} & \textbf{16.95±0.31} & 24.92±14.84 & 50.65 & 68.25 &  & CCS & 21.40±0.80 & 13.25±0.59 & 22.48±14.62 & 110.59 & 57.12 \\
 & SAPS (soft) & \textbf{26.33±0.27} & 17.08±0.30 & 28.90±17.60 & 37.01 & 72.31 &  & SAPS (soft) & 21.24±0.32 & 13.11±0.42 & \textbf{16.45±16.14} & \textbf{151.01} & \textbf{50.80} \\
 & SAPS & 26.68±0.46 & {\ul 17.03±0.45} & {\ul 22.64±17.16} & \textbf{58.46} & {\ul 66.35} &  & SAPS & 21.81±1.47 & 13.28±0.81 & {\ul 20.16±16.68} & {\ul 126.14} & 55.25 \\ \hline
\end{tabular}
}
\end{table*}

\subsection{Efficiency}
A principal motivation for employing sample selection strategies in model repair is the pursuit of efficiency. Efficiency here encompasses not only the reduction in data volume but also the overall computational cost and the value extracted from each processed sample. To capture these aspects, we evaluate both the direct time overhead of different strategies and a holistic indicator, the Repair Efficiency Score, which balances detoxification effectiveness against the extent of data reduction. The detailed results of these evaluations are reported in Tables \ref{table_timegpt} and \ref{table_timeneo}.

\subsubsection{Computational Overhead of Selection Strategies.} The total runtime of a repair pipeline can be decomposed into two components: the time required to select the subset of data and the subsequent repair time. Our analysis highlights a pronounced contrast between lightweight and computationally intensive strategies. Methods such as CCS and GraNd demand excessive resources during the selection stage. For instance, CCS requires more than 13000 seconds merely to identify samples for IRepair on GPT-2 Large, a duration that already exceeds the subsequent repair phase itself. Consequently, their overall runtime is substantially higher than the baseline of full dataset fine-tuning, thereby undermining their practical value. By contrast, lightweight strategies such as Random sampling and SAPS introduce only negligible or modest additional cost. Random sampling completes within a constant time of approximately 0.1 seconds regardless of dataset size, while SAPS requires only a few seconds, typically between one and ten. Importantly, SAPS not only adds little overhead but also reduces the subsequent repair time, leading to markedly lower total runtime. For example, the total cost of DAPT with KL constraint on GPT-2 Large is 1086 seconds with SAPS, compared to 2154 seconds when the full dataset is used. These findings suggest that computationally demanding strategies can only be justified when they deliver disproportionately large improvements in repair quality, a balance that is rarely achieved in practice.

\subsubsection{Holistic Efficiency via the Repair Efficiency Score.} To assess efficiency in a more comprehensive manner, we introduce the Repair Efficiency Score. This metric reflects how well a method balances repair effectiveness with data economy. Across diverse settings, SAPS demonstrates strong and consistent performance, achieving high RES values. For instance, in DAPT on GPT-2 Large its RES reaches 219.81, while in DAPT with KL constraint on GPT-2 XL it achieves 186.89. These results demonstrate that the semantic boundary identification mechanism in SAPS successfully isolates the most informative samples for detoxification, thereby maximizing the utility derived from each selected point. The consistency of these outcomes across both model scales and repair methods highlights the robustness of SAPS as a reliable choice for efficiency-driven repair.

\subsubsection{Relative Efficiency of Alternative Strategies.} The results also shed light on the limitations of computationally heavy methods. GraNd, despite its theoretical grounding in gradient-based importance estimation, often yields RES values well below average in the context of model repair. For example, in DAPT on GPT-Neo it produces a RES of –1.57, indicating that the marginal improvements in detoxification are insufficient to justify the large sampling overhead. Conversely, Random sampling remains an unexpectedly strong and stable baseline. Although it is frequently surpassed by SAPS, it nonetheless achieves competitive RES values in many configurations. In particular, for large models such as Pythia, Random sampling attains an RES of 226.60 in the DAPT setting. This suggests that in high-capacity models with broad representational coverage, randomly chosen subsets can already provide an adequate corrective signal, challenging the necessity of elaborate curation in all scenarios.

\subsubsection{Summary of Findings.} From an engineering standpoint, these results highlight efficiency as a multi-dimensional objective that combines computational cost with the effective use of data. Sample selection should therefore be treated as a design choice with direct implications for the feasibility of repair pipelines. Strategies with high selection overhead, such as CCS and GraNd, often prove impractical because their runtime surpasses that of full dataset repair without yielding proportionate benefits. In contrast, lightweight strategies such as Random sampling and SAPS introduce negligible or modest overhead while delivering competitive or superior outcomes. The Repair Efficiency Score confirms this balance: Random sampling can provide a strong baseline for large models or robust methods like DPO, while SAPS consistently maximizes the value of each selected sample across smaller models and methods such as DAPT or IRepair. This suggests that practitioners should match strategies to context: random subsets are appropriate when model capacity or method robustness reduces sensitivity to curation, whereas semantically informed selection is necessary when careful targeting of data yields substantial improvements. Overall, SAPS emerges as a reliable and efficient option for default use, but efficiency-focused repair should always report both runtime and RES values to ensure transparency in evaluating trade-offs.

\begin{table*}[]
\centering
\caption{Computational efficiency comparison of sample selection strategies for five repair methods on GPT-2 Large and GPT-2 XL. Reported metrics include sampling time, repair time, and the Repair Efficiency Score (RES).}
\label{table_timegpt}
\resizebox{0.85\linewidth}{!}{
\setlength{\tabcolsep}{2pt}
\begin{tabular}{cccccccccccc}
\hline
\multicolumn{6}{c}{GPT-2 Large} & \multicolumn{6}{c}{GPT-2 XL} \\ \hline
\begin{tabular}[c]{@{}c@{}}Repair \\ Method\end{tabular} & \begin{tabular}[c]{@{}c@{}}Selection \\ Strategy\end{tabular} & \begin{tabular}[c]{@{}c@{}}Sampling\\ Time (s)\end{tabular} & \begin{tabular}[c]{@{}c@{}}Repair\\ Time (s)\end{tabular} & \begin{tabular}[c]{@{}c@{}}Total\\ Time (s)\end{tabular} & \begin{tabular}[c]{@{}c@{}}RES\\ $\uparrow$\end{tabular} & \begin{tabular}[c]{@{}c@{}}Repair \\ Method\end{tabular} & \begin{tabular}[c]{@{}c@{}}Selection \\ Strategy\end{tabular} & \begin{tabular}[c]{@{}c@{}}Sampling\\ Time (s)\end{tabular} & \begin{tabular}[c]{@{}c@{}}Repair\\ Time (s)\end{tabular} & \begin{tabular}[c]{@{}c@{}}Total\\ Time (s)\end{tabular} & \begin{tabular}[c]{@{}c@{}}RES\\ $\uparrow$\end{tabular} \\ \hline
\multirow{7}{*}{DAPT} & Full Sample & — & 2149.4 & 2149.4 & — & \multirow{7}{*}{DAPT} & Full Sample & — & 3698.4 & 3698.4 & — \\
 & Random & 0.1 & 1239.6 & 1239.7 & 177.88 &  & Random & 0.1 & 2234.8 & \textbf{2234.9} & 79.05 \\
 & K-Center & 46.6 & 1368.8 & 1415.4 & {\ul 195.50} &  & K-Center & 20.5 & 2629.8 & 2650.3 & \textbf{150.44} \\
 & GraNd & 390.8 & 1234.0 & 1624.8 & 94.36 &  & GraNd & 448.0 & 2255.6 & 2703.6 & 115.84 \\
 & CCS & 5954.4 & 1336.0 & 7290.4 & 63.87 &  & CCS & 4003.2 & 2664.0 & 6667.2 & 75.60 \\
 & SAPS (soft) & 4.1 & 1227.2 & {\ul 1231.3} & 148.53 &  & SAPS (soft) & 3.3 & 2288.2 & {\ul 2291.5} & 129.24 \\
 & SAPS & 4.2 & 1220.8 & \textbf{1225.0} & \textbf{219.81} &  & SAPS & 3.3 & 2374.2 & 2377.5 & {\ul 148.32} \\ \hline
\multirow{7}{*}{DAPT+KL} & Full Sample & — & 2154.4 & 2154.4 & — & \multirow{7}{*}{DAPT+KL} & Full Sample & — & 3616.2 & 3616.2 & — \\
 & Random & 0.1 & 1292.6 & 1292.7 & 63.07 &  & Random & 0.1 & 2358.0 & 2358.1 & 102.29 \\
 & K-Center & 33.1 & 1059.4 & {\ul 1092.5} & 34.55 &  & K-Center & 22.7 & 2215.4 & 2238.1 & 40.69 \\
 & GraNd & 695.6 & 1159.4 & 1855.0 & 43.16 &  & GraNd & 476.0 & 2269.4 & 2745.4 & 77.98 \\
 & CCS & 691.3 & 1186.2 & 1877.5 & {\ul 147.08} &  & CCS & 4255.2 & 2603.0 & 6858.2 & 76.51 \\
 & SAPS (soft) & 4.9 & 1269.2 & 1274.1 & 135.51 &  & SAPS (soft) & 3.4 & 2147.2 & \textbf{2150.6} & {\ul 133.80} \\
 & SAPS & 5.2 & 1081.2 & \textbf{1086.4} & \textbf{176.75} &  & SAPS & 3.5 & 2216.2 & {\ul 2219.66} & \textbf{186.89} \\ \hline
\multirow{7}{*}{DPO} & Full Sample & — & 742.6 & 742.6 & — & \multirow{7}{*}{DPO} & Full Sample & — & 1687.2 & 1687.2 & — \\
 & Random & 0.1 & 317.8 & 317.9 & 50.63 &  & Random & 0.1 & 858.2 & 858.3 & {\ul 66.57} \\
 & K-Center & 4.20 & 286.8 & \textbf{291.0} & 47.52 &  & K-Center & 8.6 & 859.2 & 867.8 & 56.78 \\
 & GraNd & 75.8 & 611.6 & 687.4 & 47.64 &  & GraNd & 264.40 & 796.2 & 1060.6 & 27.55 \\
 & CCS & 1188.0 & 662.8 & 1850.8 & \textbf{93.90} &  & CCS & 1893.6 & 881.6 & 2775.2 & 41.89 \\
 & SAPS (soft) & 1.9 & 309.8 & {\ul 311.70} & {\ul 61.48} &  & SAPS (soft) & 2.02 & 633.0 & {\ul 635.0} & 32.72 \\
 & SAPS & 1.7 & 506.0 & 507.7 & 42.40 &  & SAPS & 2.0 & 603.6 & \textbf{605.6} & \textbf{67.29} \\ \hline
\multirow{7}{*}{IRepair} & Full Sample & — & 4484.8 & 4484.8 & — & \multirow{7}{*}{IRepair} & Full Sample & — & 3916.6 & 3916.6 & — \\
 & Random & 0.1 & 2654.8 & 2654.9 & 109.39 &  & Random & 0.1 & 2164.2 & \textbf{2164.3} & {\ul 123.42} \\
 & K-Center & 207.5 & 2434.4 & {\ul 2641.9} & \textbf{192.47} &  & K-Center & 72.8 & 2226.0 & 2298.8 & 84.23 \\
 & GraNd & 1092.9 & 2448.0 & 3540.9 & {\ul 174.39} &  & GraNd & 897.8 & 2147.8 & 3045.6 & \textbf{127.49} \\
 & CCS & 13226.4 & 2481.4 & 15707.8 & 119.35 &  & CCS & 8128.8 & 1880.4 & 10009.2 & 109.97 \\
 & SAPS (soft) & 8.4 & 3029.4 & 3037.8 & 147.52 &  & SAPS (soft) & 5.6 & 2252.6 & 2258.2 & 104.02 \\
 & SAPS & 10.3 & 2568.6 & \textbf{2578.9} & 122.70 &  & SAPS & 5.8 & 2246.2 & {\ul 2252.0} & 116.50 \\ \hline
\multirow{7}{*}{IRepair+KL} & Full Sample & — & 3132.8 & 3132.8 & — & \multirow{7}{*}{IRepair+KL} & Full Sample & — & 3738.0 & 3738.0 & — \\
 & Random & 0.1 & 1591.6 & \textbf{1591.7} & 100.54 &  & Random & 0.1 & 2495.6 & 2495.7 & 53.88 \\
 & K-Center & 73.3 & 1696.2 & 1769.5 & {\ul 106.79} &  & K-Center & 79.6 & 2358.2 & 2437.8 & {\ul 101.80} \\
 & GraNd & 505.2 & 1741.2 & 2246.4 & 67.67 &  & GraNd & 1052.8 & 2182.2 & 3235.0 & 62.01 \\
 & CCS & 7509.6 & 1924.2 & 9433.8 & 58.86 &  & CCS & 7696.8 & 2015.2 & 9712.0 & 98.67 \\
 & SAPS (soft) & 5.1 & 1698.8 & 1703.9 & 81.63 &  & SAPS (soft) & 5.3 & 2243.60 & {\ul 2248.9} & 69.68 \\
 & SAPS & 5.20 & 1692.4 & {\ul 1697.6} & \textbf{116.42} &  & SAPS & 5.3 & 2132.8 & \textbf{2138.1} & \textbf{113.09} \\ \hline
\end{tabular}}
\end{table*}

\begin{table*}[]
\centering
\caption{Computational efficiency comparison of sample selection strategies for five repair methods on GPT-Neo and Pythia. Reported metrics include sampling time, repair time, and the Repair Efficiency Score (RES).}
\label{table_timeneo}
\resizebox{0.85\linewidth}{!}{
\setlength{\tabcolsep}{2pt}
\begin{tabular}{cccccccccccc}
\hline
\multicolumn{6}{c}{GPT-Neo} & \multicolumn{6}{c}{Pythia} \\ \hline
\begin{tabular}[c]{@{}c@{}}Repair \\ Method\end{tabular} & \begin{tabular}[c]{@{}c@{}}Selection \\ Strategy\end{tabular} & \begin{tabular}[c]{@{}c@{}}Sampling\\ Time (s)\end{tabular} & \begin{tabular}[c]{@{}c@{}}Repair\\ Time (s)\end{tabular} & \begin{tabular}[c]{@{}c@{}}Total\\ Time (s)\end{tabular} & \begin{tabular}[c]{@{}c@{}}RES\\ $\uparrow$\end{tabular} & \begin{tabular}[c]{@{}c@{}}Repair \\ Method\end{tabular} & \begin{tabular}[c]{@{}c@{}}Selection \\ Strategy\end{tabular} & \begin{tabular}[c]{@{}c@{}}Sampling\\ Time (s)\end{tabular} & \begin{tabular}[c]{@{}c@{}}Repair\\ Time (s)\end{tabular} & \begin{tabular}[c]{@{}c@{}}Total\\ Time (s)\end{tabular} & \begin{tabular}[c]{@{}c@{}}RES\\ $\uparrow$\end{tabular} \\ \hline
\multirow{7}{*}{DAPT} & Full Sample & — & 2720.4 & 2720.4 & — & \multirow{7}{*}{DAPT} & Full Sample & — & 3177.6 & 3177.6 & — \\
 & Random & 0.1 & 1739.2 & 1739.3 & 18.81 &  & Random & 0.1 & 1731.6 & {\ul 1731.7} & \textbf{226.60} \\
 & K-Center & 17.5 & 1736.4 & 1753.9 & 12.176 &  & K-Center & 24.7 & 1755.0 & 1779.7 & 183.73 \\
 & GraNd & 391.4 & 1910.4 & 2301.8 & -1.57 &  & GraNd & 915.6 & 1743.8 & 2659.4 & 107.03 \\
 & CCS & 3484.8 & 1784.6 & 5269.4 & {\ul 20.62} &  & CCS & 3931.2 & 2161.0 & 6092.2 & 166.81 \\
 & SAPS (soft) & 3.1 & 1678.6 & \textbf{1681.7} & -59.96 &  & SAPS (soft) & 4.8 & 2156.4 & 2161.2 & 37.94 \\
 & SAPS & 3.0 & 1732.0 & {\ul 1735.0} & \textbf{35.13} &  & SAPS & 4.1 & 1705.0 & \textbf{1709.1} & {\ul 195.47} \\ \hline
\multirow{7}{*}{DAPT+KL} & Full Sample & — & 2847.2 & 2847.2 & — & \multirow{7}{*}{DAPT+KL} & Full Sample & — & 2144.6 & 2144.6 & — \\
 & Random & 0.1 & 1758.2 & \textbf{1758.3} & {\ul -35.75} &  & Random & 0.1 & 1065.8 & 1065.9 & \textbf{337.28} \\
 & K-Center & 32.0 & 1856.8 & 1888.8 & -51.05 &  & K-Center & 9.1 & 1036.6 & \textbf{1045.7} & 164.22 \\
 & GraNd & 594.6 & 1825.2 & 2419.8 & -78.12 &  & GraNd & 542.6 & 1041.8 & 1584.4 & 230.28 \\
 & CCS & 4485.6 & 1844.4 & 6330.0 & -55.06 &  & CCS & 2001.6 & 1309.0 & 3310.6 & 301.92 \\
 & SAPS (soft) & 3.4 & 1826.0 & {\ul 1829.4} & -63.20 &  & SAPS (soft) & 2.3 & 1156.2 & 1158.5 & 216.78 \\
 & SAPS & 3.4 & 1934.0 & 1937.4 & \textbf{-25.43} &  & SAPS & 2.6 & 1047.6 & {\ul 1050.2} & {\ul 318.66} \\ \hline
\multirow{7}{*}{DPO} & Full Sample & — & 2492.4 & 2492.4 & — & \multirow{7}{*}{DPO} & Full Sample & — & 2456.4 & 2456.4 & — \\
 & Random & 0.1 & 1296.2 & \textbf{1296.3} & 51.80 &  & Random & 0.1 & 1310.4 & {\ul 1310.5} & \textbf{105.2} \\
 & K-Center & 13.2 & 1614.2 & 1627.4 & 69.10 &  & K-Center & 6.1 & 1306.0 & 1312.1 & {\ul 103.0} \\
 & GraNd & 248.6 & 1519.0 & 1767.6 & {\ul 73.03} &  & GraNd & 175.6 & 1051.2 & \textbf{1226.8} & 102.5 \\
 & CCS & 2102.4 & 1334.2 & 3436.6 & 66.21 &  & CCS & 1281.6 & 1145.2 & 2426.8 & 101.2 \\
 & SAPS (soft) & 2.1 & 1622.0 & 1624.1 & \textbf{75.56} &  & SAPS (soft) & 1.7 & 1381.0 & 1382.7 & 102.2 \\
 & SAPS & 2.2 & 1355.4 & {\ul 1357.6} & 60.19 &  & SAPS & 1.8 & 1457.6 & 1459.4 & 102.0 \\ \hline
\multirow{7}{*}{IRepair} & Full Sample & — & 5383.2 & 5383.2 & — & \multirow{7}{*}{IRepair} & Full Sample & — & 4480.2 & 4480.2 & — \\
 & Random & 0.1 & 2650.2 & {\ul 2650.3} & {\ul 165.62} &  & Random & 0.1 & 2019.2 & 2019.3 & {\ul 115.68} \\
 & K-Center & 125.94 & 2611.2 & 2737.1 & 108.09 &  & K-Center & 53.0 & 1617.6 & \textbf{1670.6} & 75.66 \\
 & GraNd & 2569.6 & 2895.8 & 5465.4 & 122.99 &  & GraNd & 1944.8 & 2244.6 & 4189.4 & 110.92 \\
 & CCS & 10411.2 & 2500.6 & 12911.8 & 180.82 &  & CCS & 6120.0 & 1324.6 & 7444.6 & 51.25 \\
 & SAPS (soft) & 7.64 & 2689.8 & 2697.4 & \textbf{171.69} &  & SAPS (soft) & 4.5 & 2060.4 & 2064.9 & \textbf{141.49} \\
 & SAPS & 8.86 & 2355.2 & \textbf{2364.1} & 131.83 &  & SAPS & 4.5 & 1907.4 & {\ul 1911.9} & 85.89 \\ \hline
\multirow{7}{*}{IRepair+KL} & Full Sample & — & 3588.0 & 3588.0 & — & \multirow{7}{*}{IRepair+KL} & Full Sample & — & 2386.2 & 2386.2 & — \\
 & Random & 0.1 & 1900.2 & 1900.3 & 59.95 &  & Random & 0.1 & 577.8 & \textbf{577.9} & 165.12 \\
 & K-Center & 42.0 & 1850.0 & 1892.0 & {\ul 80.31} &  & K-Center & 11.9 & 636.8 & 648.7 & 125.60 \\
 & GraNd & 658.8 & 1883.8 & 2542.6 & 13.92 &  & GraNd & 602.0 & 773.4 & 1375.4 & 174.03 \\
 & CCS & 5889.6 & 1554.6 & 7444.2 & 71.63 &  & CCS & 2520.0 & 651.2 & 3171.2 & 156.40 \\
 & SAPS (soft) & 5.1 & 1827.6 & \textbf{1832.7} & 52.34 &  & SAPS (soft) & 2.4 & 663.0 & 665.4 & \textbf{213.56} \\
 & SAPS & 4.9 & 1884.2 & {\ul 1889.1} & \textbf{82.67} &  & SAPS & 2.8 & 638.4 & {\ul 641.2} & {\ul 178.39} \\ \hline
\end{tabular}
}
\end{table*}

\subsection{Scaling}

\begin{figure} [ht]
	\centering  
        \subfloat[GPT-2 XL]{\includegraphics[width=1.0\linewidth]{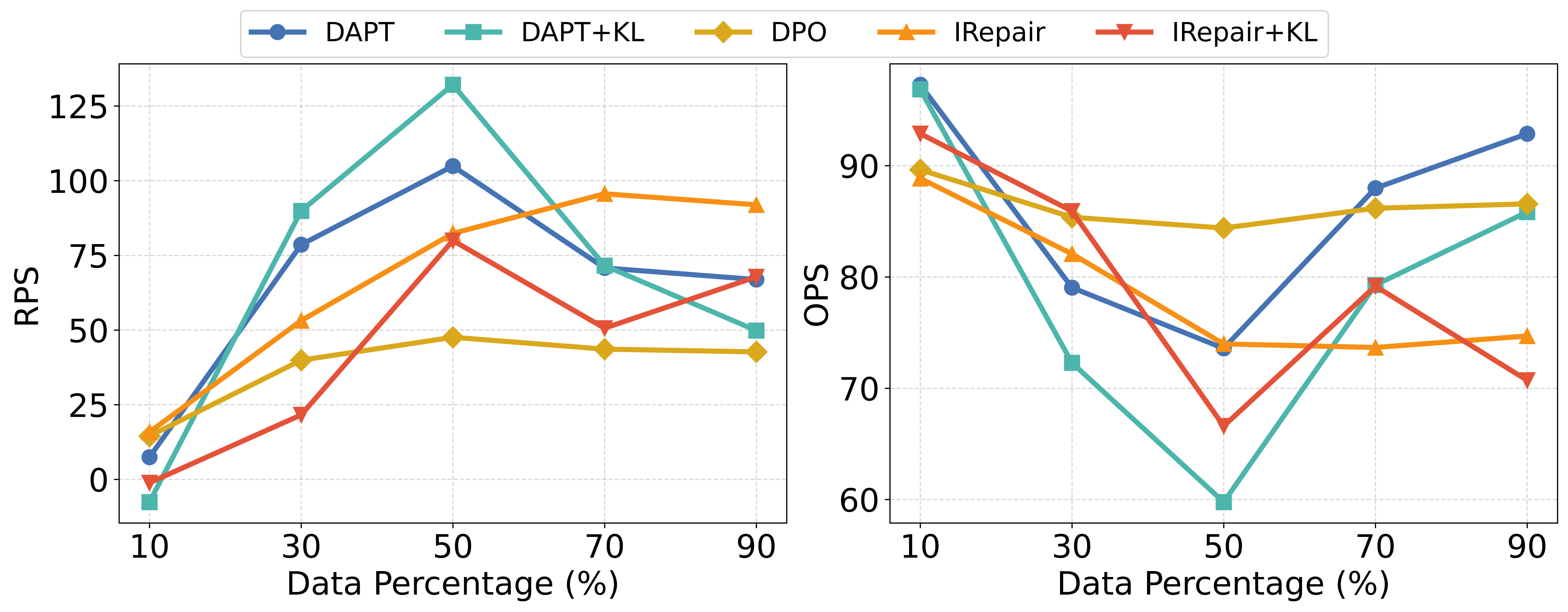}%
        \label{sub.1}}\\
        \subfloat[Pythia]{\includegraphics[width=1.0\linewidth]{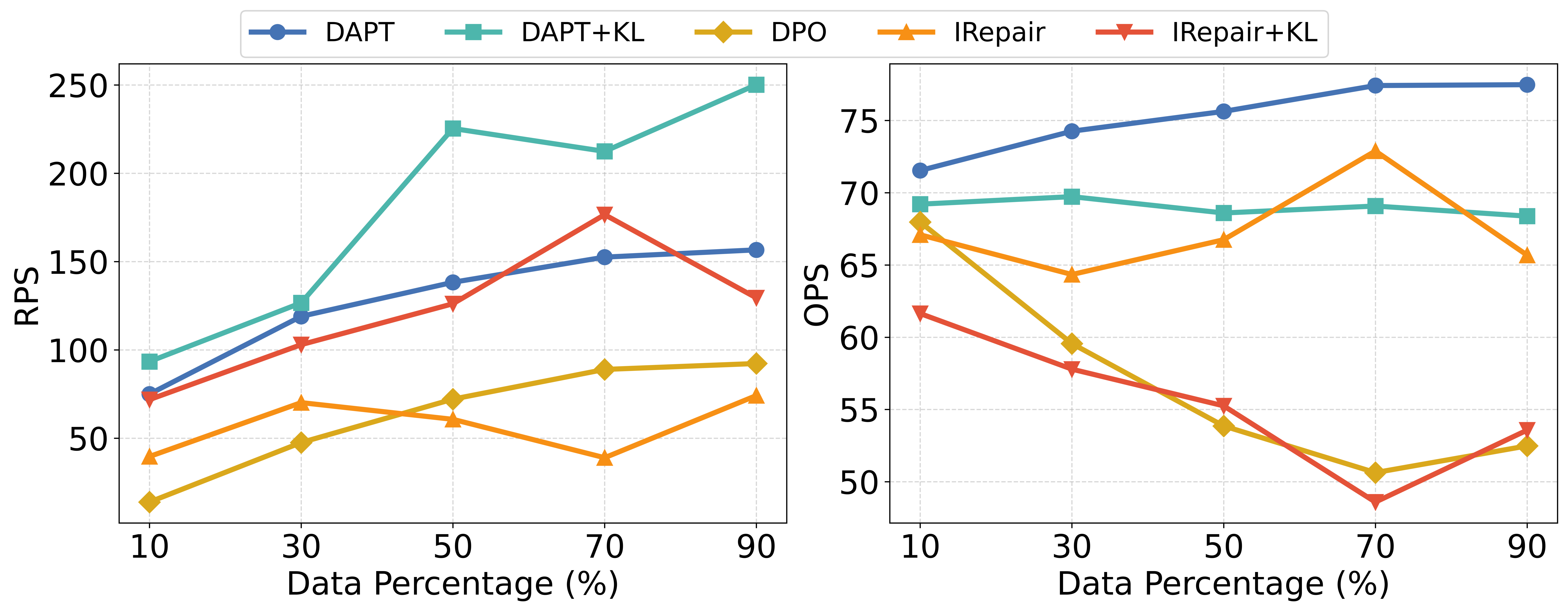}%
        \label{sub.2}}

	\caption{Comparison of five repair methods under different data selection percentages using our selection strategy. 
The left figure reports the Repair Proximity Score (RPS) relative to the full dataset baseline, while the right figure shows the Overall Performance Score (OPS). Both metrics are evaluated across data percentages ranging from 10\% to 90\%.}
	\label{percentage}
\end{figure}

A principal motivation for employing the SAPS selection strategy is to achieve effective model repair with a minimal subset of data. However, the optimal size of this subset remains an open question. To investigate this, we evaluate how the performance of SAPS, measured by both the RPS and the OPS, scales with the proportion of data used for repair. The experiments are conducted on two representative models: GPT-2 XL and the larger Pythia model, across five repair methods. The detailed results are reported in Figures~\ref{percentage} and \ref{percentage_time}.

\subsubsection{Identification of Performance Saturation.} The relationship between data proportion and performance is not linear. For certain method and model configurations, we identify a clear performance saturation point beyond which additional data yields diminishing returns. This is particularly evident with the GPT-2 XL model under the DAPT with KL constraint repair method. Here, the RPS peaks at the value of 132.15 using only fifty percent of the data. Increasing the data proportion to 70 and 90 percent sees the score drop to 71.56 and 49.82, respectively. This indicates that SAPS is highly effective at identifying a critical and informative subset of data beyond which additional samples may introduce redundancy or noise, thereby reducing the efficiency of the repair process. A similar saturation trend is observed for the IRepair with KL constraint on GPT-2 XL, where the highest score is achieved at fifty percent data.

\subsubsection{Method-Specific Sensitivity to Data Volume.} 
\begin{figure} [ht]
	\centering  
        \subfloat[GPT-2 XL]{\includegraphics[width=0.5\linewidth]{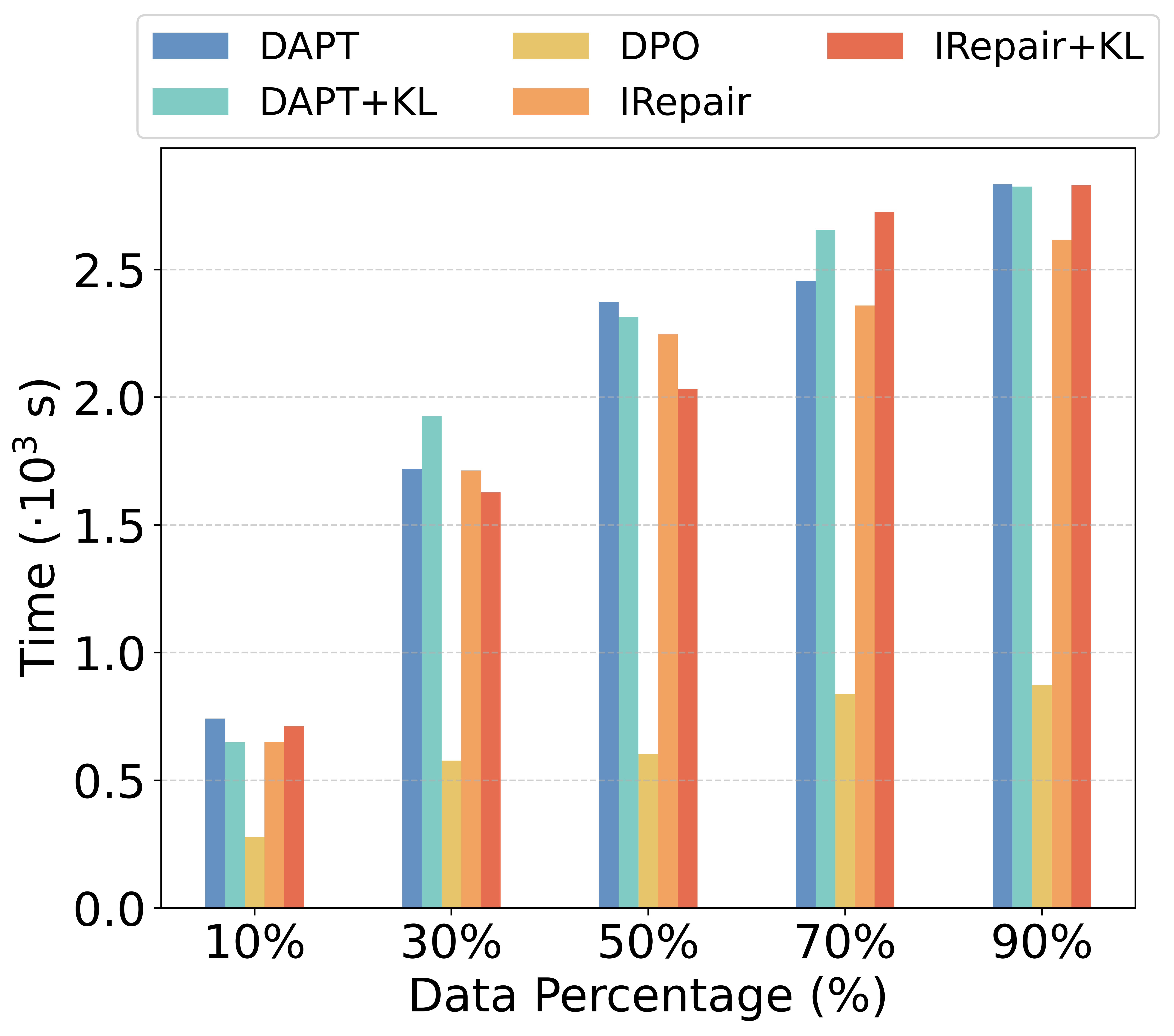}%
        \label{sub1.1}}
        \subfloat[Pythia]{\includegraphics[width=0.5\linewidth]{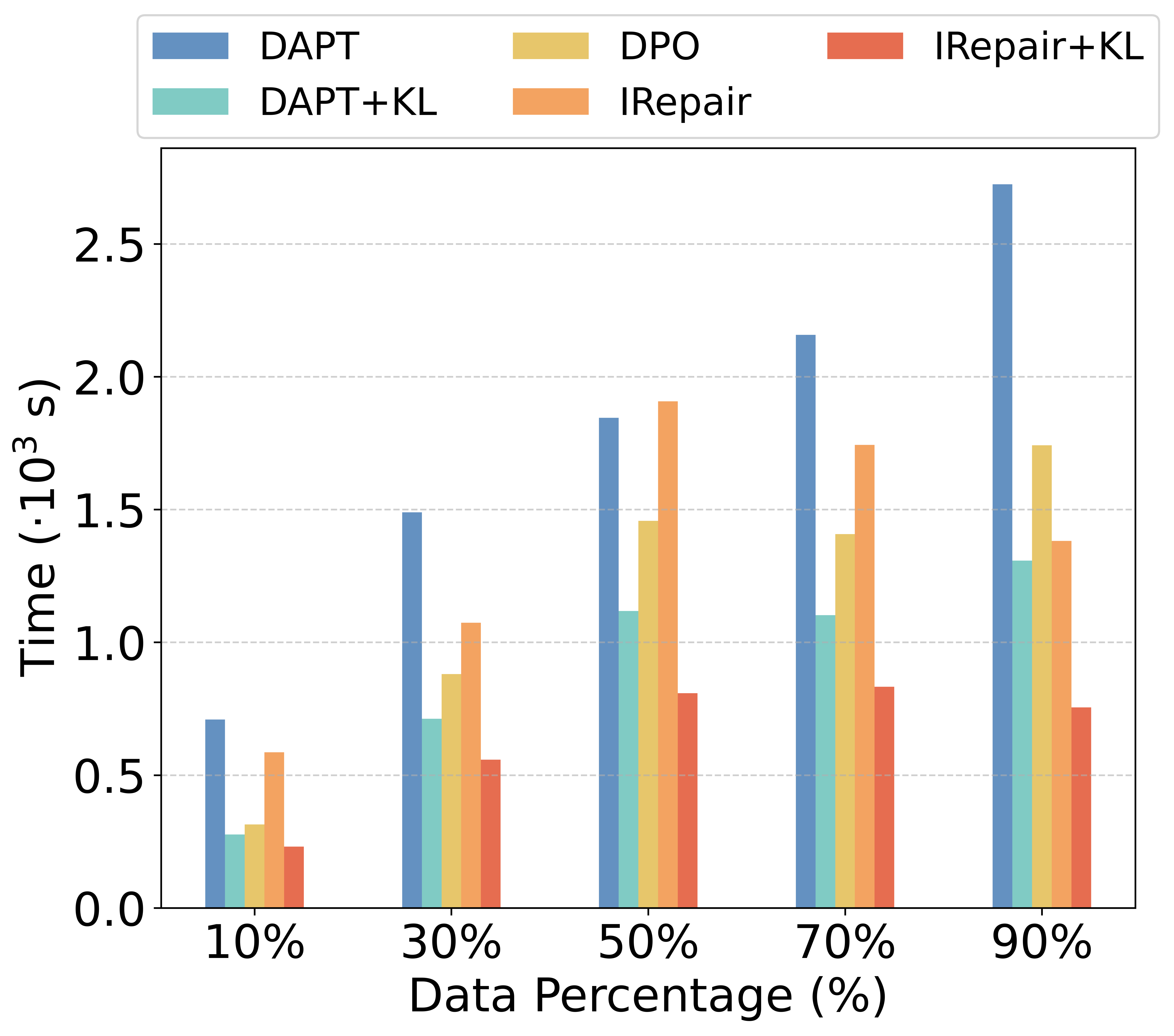}%
        \label{sub1.2}}

	\caption{Comparison of five repair methods under different data selection percentages using our selection strategy. The figure reports the repair time evaluated across data percentages ranging from 10\% to 90\%.}
	\label{percentage_time}
\end{figure}

The efficacy of varying data proportions is highly dependent on the repair method. The DPO method demonstrates considerable robustness to data quantity once a minimal threshold is met. On GPT-2 XL, its RPS rises from 14.48 at 10\% data to a stable plateau, fluctuating within a range of 39.91 to 47.57 across proportions from 30\% to 90\%. This indicates that while a minimal amount of data is necessary to initiate learning, DPO's performance quickly stabilizes, reinforcing the conclusion that it derives its learning signal primarily from the quality of preference comparisons rather than the absolute volume of data. In contrast, methods like IRepair and IRepair with KL constraint exhibit significant volatility across the entire spectrum. For example, the RPS for IRepair with KL constraint on GPT-2 XL jumps from -1.15 at 10\% data to 79.96 at 50\%, before falling to 67.80 at 90\%. This suggests that these methods are highly sensitive to both the quantity and the specific composition of the data, requiring a precise ``Goldilocks'' zone to perform optimally, a target that the SAPS strategy is designed to achieve. Notably, repair methods with higher data sensitivity also show steeper time growth curves in Figure~\ref{percentage_time}, reflecting their greater computational dependence on dataset size.

\subsubsection{Superior Data Utilization in Larger Models.} The scaling behavior is further influenced by model scale. The larger Pythia model consistently demonstrates a superior ability to leverage increased data volumes provided by SAPS compared to GPT-2 XL. This is unmistakably clear in the continuous improvement of RPS for DAPT and DAPT with KL constraint on Pythia. The RPS for DAPT with KL constraint climbs from 93.40 at 10\% data to 250.14 at 90\% data, while its OPS remains stable, even slightly improving. This indicates that larger model capacities enable more effective distillation of information from the larger, high-quality subsets curated by SAPS, translating into greater repair efficiency without degrading overall utility.

\subsubsection{Summary of Findings.} From an engineering perspective, determining the proportion of data for SAPS is a critical tuning parameter that balances speed against performance. There is no universal optimal proportion as it is contingent upon the specific model and method. For the GPT-2 XL model, a moderate proportion of 30\% to 50\% is often sufficient to reach peak performance for many methods, making it the most efficient operating point. For the larger Pythia model, a higher data proportion of 70\% to 90\% is generally beneficial, as its greater capacity allows it to convert more data into better performance. The DPO method is a notable exception, being largely insensitive to data volume, which allows for the use of minimal data subsets to maximize speed. Therefore, practitioners should calibrate the data proportion based on the target model and repair method, opting for lower proportions for faster iteration or with robust methods like DPO, and selecting higher proportions to maximize the performance of larger models or more data-sensitive methods such as DAPT with KL constraint.

\subsection{Boundary Value}
\begin{figure*} [ht]
	\centering  
        \subfloat[GPT-2 XL]{\includegraphics[width=1.0\linewidth]{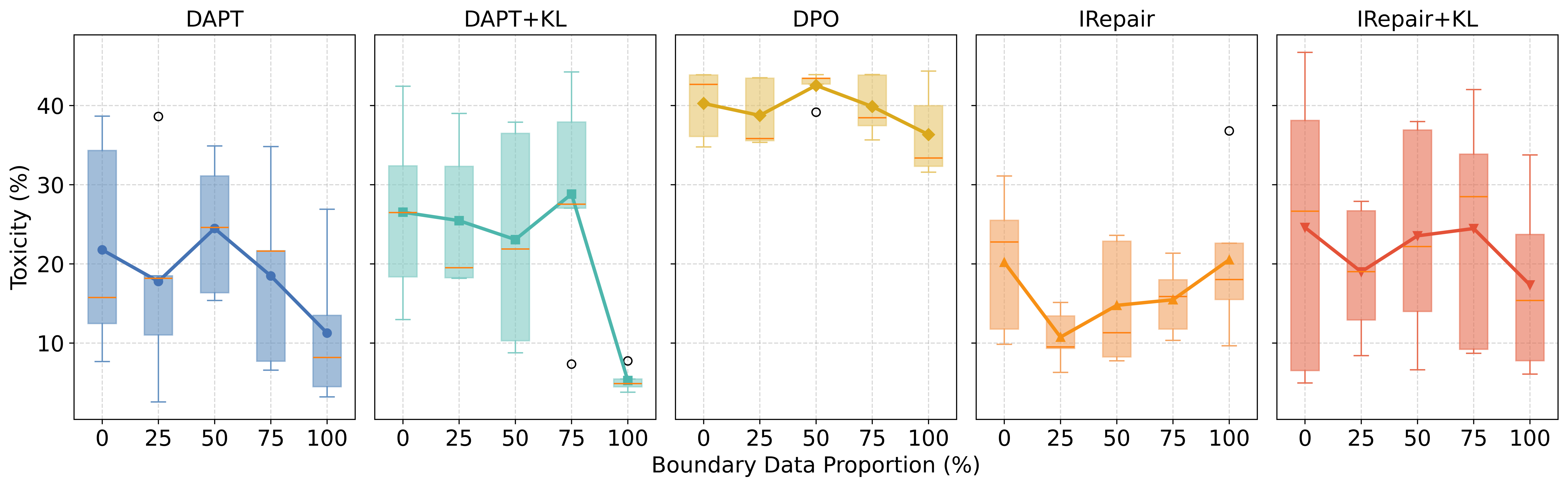}%
        \label{sub4.1}}\\
        \subfloat[Pythia]{\includegraphics[width=1.0\linewidth]{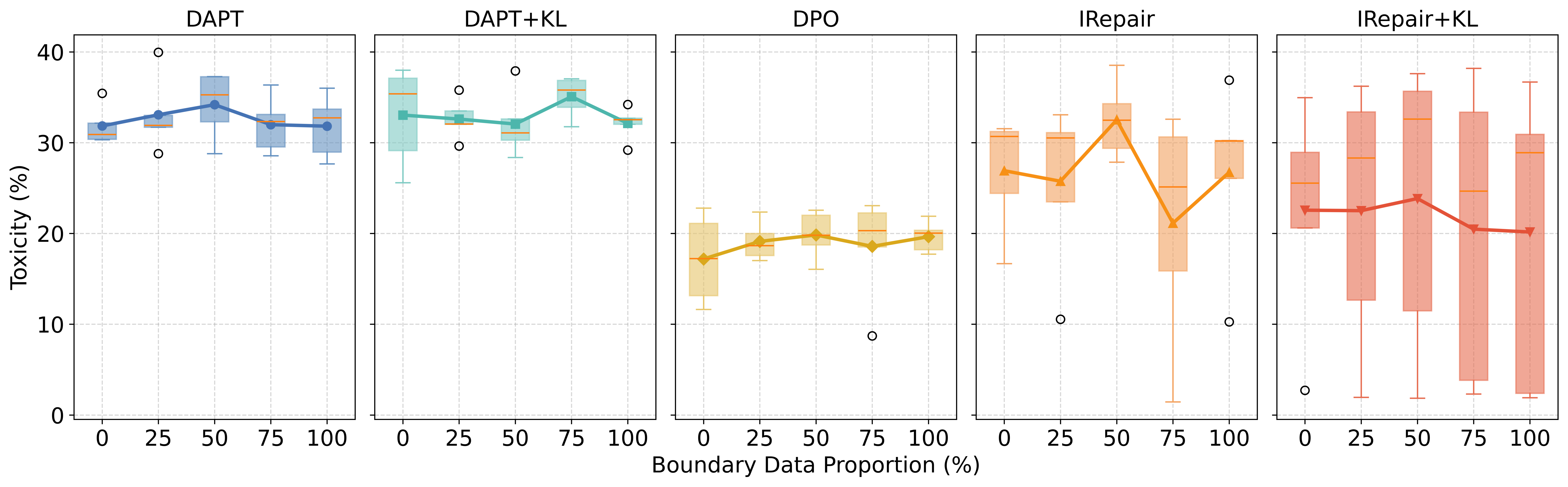}%
        \label{sub4.2}}

	\caption{Comparison of toxicity changes with varying proportions of boundary data selected by SAPS across repair methods.}
	\label{gpt2xl_boundary}
\end{figure*}

We assess the value of boundary data by applying various repair methods to both the GPT-2 XL and the larger Pythia-2.8B models, using the boundary-to-center data ratio defined by the SAPS clustering strategy. Results are shown in Figure \ref{gpt2xl_boundary}.


For DAPT on GPT-2 XL, toxicity decreases from 21.78 with no boundary samples to 11.27 when the dataset consists entirely of boundary points, indicating that boundary information is highly valuable for this class of continued pretraining methods. DAPT with KL constraint shows an even stronger dependence, with toxicity falling from 26.53 to 5.28 as the proportion of boundary data rises to 100\%. DPO, by contrast, remains largely stable across proportions, with toxicity varying only between 40.27 and 36.33, although the lowest value still occurs when boundary samples dominate. IRepair differs from these trends. Its optimal outcome emerges when only a quarter of the data are boundary samples, producing the lowest toxicity of 10.75, while performance deteriorates when the boundary share increases further. However, when KL regularization is introduced, IRepair with KL constraint aligns more closely with DAPT: toxicity steadily declines with higher boundary proportions, reaching 17.35 at full boundary selection. However, this clear pattern diminishes in the larger Pythia-2.8B model. The overall influence of the boundary-to-center ratio is less pronounced, corroborating the finding in Section \ref{compatibility} that larger models are less sensitive to nuanced sample selection. Despite this, the value of boundary data is still evident in specific methods: IRepair with KL regularization and DAPT with KL regularization exhibit their lowest toxicity scores at high boundary proportions, indicating that the core mechanism of boundary-sample-based repair remains valid but is modulated by model capacity.


From an engineering standpoint, SAPS proves to be an effective selection strategy, but its deployment must be calibrated to both the repair method and the model scale. For GPT-2 XL and similar-scale models, boundary-focused sampling is clearly advantageous for DAPT and its regularized variant. For the larger Pythia model, while the absolute gains are smaller, prioritizing boundary samples remains beneficial for KL-regularized methods. These findings suggest that as model capacity increases, the necessity for sophisticated selection may decrease, but the strategic selection of challenging samples continues to provide a consistent, if more modest, improvement in repair efficacy.

\section{Discussion}
\subsection{Generality of Our Study}
Generality Across Model Scales. Our conclusions are not confined to a single model scale. The study covers models ranging from approximately 774 million parameters to 2.8 billion parameters. Key findings, such as the superiority of the SAPS strategy in most cases and the relative insensitivity of DPO to data selection, are consistently validated across these models. This suggests that our conclusions may extend to a broader family of pretrained language models with similar architectures. The observation that larger models benefit more from increased data also resonates with established understanding in scaling laws, reinforcing the generality of our results across different resource settings.

Generality Across Repair Paradigms. The conclusions of this study hold across multiple repair paradigms. We examine five representative approaches, namely continued pretraining, continued pretraining with KL regularization, preference optimization, and parameter-localized repair. Although these methods differ fundamentally in their mechanisms, the importance of data selection is evident across all of them, even though the degree of sensitivity and the optimal strategy varied. This provides strong evidence that data selection constitutes a core dimension in model repair, one that is largely independent of the choice of algorithm. As a result, carefully designed data selection strategies are likely to remain an essential lever for improving the efficiency and effectiveness of future repair methods.

Robustness Across Evaluation Dimensions. Finally, the robustness of our findings is demonstrated across multiple evaluation dimensions. Rather than relying on a single metric, such as toxicity score, we adopt a comprehensive evaluation framework that considered effectiveness, knowledge retention, efficiency, and overall trade-offs. The advantages of the SAPS strategy are established within this multidimensional evaluation, rather than being tied to a specific performance indicator. This strengthens the credibility of our conclusions and underscores that our recommendations are grounded in a balanced assessment, thereby avoiding biases that may arise from optimizing for only one dimension.

\subsection{Future Research Directions}
Automated Strategy Selection. One promising avenue for future work is the development of an automated framework for strategy selection. Our findings indicate that the optimal choice depends on the model characteristics, the repair method, and the evaluation target. A framework that can automatically recommend or even combine appropriate strategies based on model scale, architecture, repair objectives, and computational budget would significantly lower the barrier for practical deployment.

Hybrid Selection Strategies. A second direction is the design of hybrid selection strategies. While individual strategies such as representativeness-based or uncertainty-based selection each have their strengths, their combination may provide complementary benefits. For instance, a hybrid approach might first ensure dataset representativeness and then prioritize samples that are most informative for model repair. Such approaches may achieve greater effectiveness than relying on a single principle.

Extension to Other Safety and Alignment Tasks. A third direction is to extend the evaluation framework to other safety and alignment tasks. Our current study focused on detoxification. Applying the same framework to domains such as reducing social bias, improving factual accuracy, or improving the capability to follow instruction would provide valuable information. Validating the generality of these strategies across a broader set of tasks would greatly enhance their applicability and significance for the community.

\section{Threats to Validity}
This section discusses potential threats to the validity of our experimental findings and the measures taken to mitigate them, in line with established practices in empirical software engineering research.

\subsection{Internal Validity}

Threats to internal validity concern the causal relationship between the interventions, namely the sample selection strategies, and the observed outcomes in terms of repair effectiveness.

Interaction with Randomness. The performance of both selection strategies and repair algorithms may be affected by random factors such as parameter initialization, data shuffling, and stochasticity in training. We control randomness by fixing random seeds where possible, but certain stochastic elements are inherent to the process. To mitigate this, we conduct multiple runs with different seeds for key experiments and report the corresponding standard deviations in our tables, thereby enhancing the robustness of our conclusions.

Hyperparameter Configuration. All repair methods and selection strategies require setting hyperparameters such as learning rates and the number of training steps. We adopt standard configurations from prior literature and perform preliminary tuning to ensure fairness. However, it is possible that alternative strategy-method combinations yield different outcomes under other hyperparameter choices. Achieving optimal performance may require joint optimization of repair and selection parameters, which remains an open direction for further refinement.

\subsection{External Validity} 

Threats to external validity pertain to the generalizability of our findings beyond the specific conditions of this study.

Model Architectures and Scales. Our experiments are conducted on decoder-only autoregressive models belonging to the GPT-2, GPT-Neo, and Pythia families. These models represent important classes of large language models, but the conclusions may not directly extend to other architectures such as encoder-decoder models, or to extremely large-scale models with hundreds of billions of parameters. The effectiveness of selection strategies may depend on both model size and architectural characteristics.

Scope of Repair Tasks. The present study focuses on toxicity reduction as a representative repair task. Although toxicity is a central concern in model safety, the extent to which our findings apply to other repair objectives, such as mitigating social biases, improving factual accuracy, or strengthening robustness against adversarial fine-tuning-requires further investigation. Different types of undesired behavior may call for distinct principles of data selection.

Dataset Representativeness. The conclusions of this study are based on specific datasets used for both repair and evaluation. For instance, WikiText-2 is used to assess perplexity, while a dedicated benchmark is employed to measure toxicity. The performance of selection strategies may vary in domains with substantially different linguistic characteristics, such as scientific text or source code, and thus further validation is needed in diverse application contexts.

\subsection{Construct Validity} 

Threats to construct validity concern the degree to which our evaluation metrics accurately capture the intended theoretical concepts.

Toxicity Measurement. Our study relies on the Perspective API, a widely recognized tool for toxicity classification, to quantify repair success. While this provides a scalable and reproducible metric, it introduces two specific threats. First, the metric represents a single model's judgment, which may not fully align with human perception or capture nuanced, context-dependent forms of harm. Second, as a service subject to version updates, the API's scoring criteria could shift over time, potentially introducing subtle inconsistencies in measurements if experiments are replicated across different versions. Although we used a fixed version for consistency, this reliance means our toxicity scores are inherently tied to the Perspective model's specific definitions and sensitivities.

Comprehensiveness of Efficiency Metrics. RES is designed as a composite measure that balances repair effectiveness with data usage. Although this provides a holistic comparison, it may obscure the specific trade-offs among its components. To address this limitation, we report disaggregated metrics including RPS, OPS, and time cost alongside RES to give a more complete view of performance.

Perplexity as a Proxy for Language Ability. Perplexity on standard benchmarks is used as a proxy for the model’s general language proficiency. Although widely adopted, perplexity is not a perfect measure and does not always correlate with downstream task performance. Strengthening this dimension of validity would require direct evaluation on a broader suite of downstream tasks, which we identify as a promising direction for future work.

\section{Conclusion}
Our work establishes that sample selection is a critical determinant of efficacy and efficiency in model repair, on par with the choice of repair algorithm itself. Through a comprehensive empirical study, we demonstrate that semantics-based strategies, particularly SAPS, consistently achieve superior detoxification with significantly reduced data usage by identifying high-value samples. We further reveal that the optimal strategy is context-dependent: while SAPS excels for most continued pre-training methods, its advantage diminishes for DPO, and larger models exhibit a greater capacity to benefit from larger curated subsets. Our findings advocate for a paradigm shift towards data-centric repair pipelines, where intelligent sample selection is a fundamental component for scalable and responsible LLM development. To support future research and ensure reproducibility, the implementation code is publicly available at \url{https://github.com/RitaLi1005/SAPS}.

\bibliographystyle{IEEEtran}
\bibliography{ref}

\begin{thebibliography}{10}
\providecommand{\url}[1]{#1}
\csname url@samestyle\endcsname
\providecommand{\newblock}{\relax}
\providecommand{\bibinfo}[2]{#2}
\providecommand{\BIBentrySTDinterwordspacing}{\spaceskip=0pt\relax}
\providecommand{\BIBentryALTinterwordstretchfactor}{4}
\providecommand{\BIBentryALTinterwordspacing}{\spaceskip=\fontdimen2\font plus
\BIBentryALTinterwordstretchfactor\fontdimen3\font minus \fontdimen4\font\relax}
\providecommand{\BIBforeignlanguage}[2]{{%
\expandafter\ifx\csname l@#1\endcsname\relax
\typeout{** WARNING: IEEEtran.bst: No hyphenation pattern has been}%
\typeout{** loaded for the language `#1'. Using the pattern for}%
\typeout{** the default language instead.}%
\else
\language=\csname l@#1\endcsname
\fi
#2}}
\providecommand{\BIBdecl}{\relax}
\BIBdecl

\bibitem{laban2023summedits}
P.~Laban, W.~Kry{\'s}ci{\'n}ski, D.~Agarwal, A.~R. Fabbri, C.~Xiong, S.~Joty, and C.-S. Wu, ``Summedits: Measuring llm ability at factual reasoning through the lens of summarization,'' in \emph{Proceedings of the 2023 conference on empirical methods in natural language processing}, 2023, pp. 9662--9676.

\bibitem{zhang2025systematic}
H.~Zhang, P.~S. Yu, and J.~Zhang, ``A systematic survey of text summarization: From statistical methods to large language models,'' \emph{ACM Computing Surveys}, vol.~57, no.~11, pp. 1--41, 2025.

\bibitem{liu2024speak}
C.~Liu, Z.~Xie, S.~Zhao, J.~Zhou, T.~Xu, M.~Li, and E.~Chen, ``Speak from heart: an emotion-guided llm-based multimodal method for emotional dialogue generation,'' in \emph{Proceedings of the 2024 International Conference on Multimedia Retrieval}, 2024, pp. 533--542.

\bibitem{voria2025recover}
G.~Voria, F.~Casillo, C.~Gravino, G.~Catolino, and F.~Palomba, ``Recover: Toward requirements generation from stakeholders’ conversations,'' \emph{IEEE Transactions on Software Engineering}, 2025.

\bibitem{kolla2024llm}
M.~Kolla, S.~Salunkhe, E.~Chandrasekharan, and K.~Saha, ``Llm-mod: Can large language models assist content moderation?'' in \emph{Extended Abstracts of the CHI Conference on Human Factors in Computing Systems}, 2024, pp. 1--8.

\bibitem{yao2023llm}
J.-Y. Yao, K.-P. Ning, Z.-H. Liu, M.-N. Ning, Y.-Y. Liu, and L.~Yuan, ``Llm lies: Hallucinations are not bugs, but features as adversarial examples,'' \emph{arXiv preprint arXiv:2310.01469}, 2023.

\bibitem{schwarzschild2024rethinking}
A.~Schwarzschild, Z.~Feng, P.~Maini, Z.~Lipton, and J.~Z. Kolter, ``Rethinking llm memorization through the lens of adversarial compression,'' \emph{Advances in Neural Information Processing Systems}, vol.~37, pp. 56\,244--56\,267, 2024.

\bibitem{yeh2023evaluating}
K.-C. Yeh, J.-A. Chi, D.-C. Lian, and S.-K. Hsieh, ``Evaluating interfaced llm bias,'' in \emph{Proceedings of the 35th Conference on Computational Linguistics and Speech Processing (ROCLING 2023)}, 2023, pp. 292--299.

\bibitem{zhang2025challenges}
B.~Zhang and G.-J. Ren, ``Challenges and remedies of domain-specific classifiers as llm guardrails: Self-harm as a case study,'' in \emph{Proceedings of the 2025 Conference of the Nations of the Americas Chapter of the Association for Computational Linguistics: Human Language Technologies (Volume 3: Industry Track)}, 2025, pp. 173--182.

\bibitem{liu2024empirical}
M.~Liu, J.~Wang, T.~Lin, Q.~Ma, Z.~Fang, and Y.~Wu, ``An empirical study of the code generation of safety-critical software using llms.'' \emph{Applied Sciences (2076-3417)}, vol.~14, no.~3, 2024.

\bibitem{yu2024fight}
X.~Yu, L.~Liu, X.~Hu, J.~W. Keung, J.~Liu, and X.~Xia, ``Fight fire with fire: How much can we trust chatgpt on source code-related tasks?'' \emph{IEEE Transactions on Software Engineering}, 2024.

\bibitem{fatima2024flakyfix}
S.~Fatima, H.~Hemmati, and L.~C. Briand, ``Flakyfix: Using large language models for predicting flaky test fix categories and test code repair,'' \emph{IEEE Transactions on Software Engineering}, vol.~50, no.~12, pp. 3146--3171, 2024.

\bibitem{qin2025s}
Y.~Qin, S.~Wang, Y.~Lou, J.~Dong, K.~Wang, X.~Li, and X.~Mao, ``S oap fl: A standard operating procedure for llm-based method-level fault localization,'' \emph{IEEE Transactions on Software Engineering}, 2025.

\bibitem{korbak2023pretraining}
T.~Korbak, K.~Shi, A.~Chen, R.~V. Bhalerao, C.~Buckley, J.~Phang, S.~R. Bowman, and E.~Perez, ``Pretraining language models with human preferences,'' in \emph{International Conference on Machine Learning}.\hskip 1em plus 0.5em minus 0.4em\relax PMLR, 2023, pp. 17\,506--17\,533.

\bibitem{thangarasa2023spdf}
V.~Thangarasa, A.~Gupta, W.~Marshall, T.~Li, K.~Leong, D.~DeCoste, S.~Lie, and S.~Saxena, ``Spdf: Sparse pre-training and dense fine-tuning for large language models,'' in \emph{Uncertainty in Artificial Intelligence}.\hskip 1em plus 0.5em minus 0.4em\relax PMLR, 2023, pp. 2134--2146.

\bibitem{liu2023exposing}
B.~Liu, J.~Ash, S.~Goel, A.~Krishnamurthy, and C.~Zhang, ``Exposing attention glitches with flip-flop language modeling,'' \emph{Advances in Neural Information Processing Systems}, vol.~36, pp. 25\,549--25\,583, 2023.

\bibitem{chakraborty2024transfer}
S.~Chakraborty, S.~S. Ghosal, M.~Yin, D.~Manocha, M.~Wang, A.~S. Bedi, and F.~Huang, ``Transfer q-star: Principled decoding for llm alignment,'' \emph{Advances in Neural Information Processing Systems}, vol.~37, pp. 101\,725--101\,761, 2024.

\bibitem{bao2024decoding}
K.~Bao, J.~Zhang, Y.~Zhang, X.~Huo, C.~Chen, and F.~Feng, ``Decoding matters: Addressing amplification bias and homogeneity issue for llm-based recommendation,'' \emph{arXiv preprint arXiv:2406.14900}, 2024.

\bibitem{ma2025dressing}
\BIBentryALTinterwordspacing
X.~Ma, Y.~Xu, Y.~Lin, T.~Wang, X.~Chu, X.~Gao, J.~Zhao, and Y.~Wang, ``{DRESS}ing up {LLM}: Efficient stylized question-answering via style subspace editing,'' in \emph{The Thirteenth International Conference on Learning Representations}, 2025. [Online]. Available: \url{https://openreview.net/forum?id=mNVR9jJYqK}
\BIBentrySTDinterwordspacing

\bibitem{Dathathri2020Plug}
\BIBentryALTinterwordspacing
S.~Dathathri, A.~Madotto, J.~Lan, J.~Hung, E.~Frank, P.~Molino, J.~Yosinski, and R.~Liu, ``Plug and play language models: A simple approach to controlled text generation,'' in \emph{International Conference on Learning Representations}, 2020. [Online]. Available: \url{https://openreview.net/forum?id=H1edEyBKDS}
\BIBentrySTDinterwordspacing

\bibitem{gehman2020realtoxicityprompts}
S.~Gehman, S.~Gururangan, M.~Sap, Y.~Choi, and N.~A. Smith, ``Realtoxicityprompts: Evaluating neural toxic degeneration in language models,'' \emph{Findings of the Association for Computational Linguistics: EMNLP 2020}, 2020.

\bibitem{rafailov2023direct}
R.~Rafailov, A.~Sharma, E.~Mitchell, C.~D. Manning, S.~Ermon, and C.~Finn, ``Direct preference optimization: Your language model is secretly a reward model,'' \emph{Advances in Neural Information Processing Systems}, vol.~36, pp. 53\,728--53\,741, 2023.

\bibitem{wang2022exploring}
B.~Wang, W.~Ping, C.~Xiao, P.~Xu, M.~Patwary, M.~Shoeybi, B.~Li, A.~Anandkumar, and B.~Catanzaro, ``Exploring the limits of domain-adaptive training for detoxifying large-scale language models,'' \emph{Advances in Neural Information Processing Systems}, vol.~35, pp. 35\,811--35\,824, 2022.

\bibitem{zhang2024towards}
J.~Zhang, S.~Vahidian, M.~Kuo, C.~Li, R.~Zhang, T.~Yu, G.~Wang, and Y.~Chen, ``Towards building the federatedgpt: Federated instruction tuning,'' in \emph{ICASSP 2024-2024 IEEE International Conference on Acoustics, Speech and Signal Processing (ICASSP)}.\hskip 1em plus 0.5em minus 0.4em\relax IEEE, 2024, pp. 6915--6919.

\bibitem{tang2024graphgpt}
J.~Tang, Y.~Yang, W.~Wei, L.~Shi, L.~Su, S.~Cheng, D.~Yin, and C.~Huang, ``Graphgpt: Graph instruction tuning for large language models,'' in \emph{Proceedings of the 47th International ACM SIGIR Conference on Research and Development in Information Retrieval}, 2024, pp. 491--500.

\bibitem{zhang2025gpt4roi}
S.~Zhang, P.~Sun, S.~Chen, M.~Xiao, W.~Shao, W.~Zhang, Y.~Liu, K.~Chen, and P.~Luo, ``Gpt4roi: Instruction tuning large language model on region-of-interest,'' in \emph{European Conference on Computer Vision}.\hskip 1em plus 0.5em minus 0.4em\relax Springer, 2025, pp. 52--70.

\bibitem{ouyang2022training}
L.~Ouyang, J.~Wu, X.~Jiang, D.~Almeida, C.~Wainwright, P.~Mishkin, C.~Zhang, S.~Agarwal, K.~Slama, A.~Ray \emph{et~al.}, ``Training language models to follow instructions with human feedback,'' \emph{Advances in neural information processing systems}, vol.~35, pp. 27\,730--27\,744, 2022.

\bibitem{sotoudeh2019correcting}
M.~Sotoudeh and A.~Thakur, ``Correcting deep neural networks with small, generalizing patches,'' in \emph{Workshop on safety and robustness in decision making}, 2019.

\bibitem{usman2021nn}
M.~Usman, D.~Gopinath, Y.~Sun, Y.~Noller, and C.~S. P{\u{a}}s{\u{a}}reanu, ``Nn repair: Constraint-based repair of neural network classifiers,'' in \emph{Computer Aided Verification: 33rd International Conference, CAV 2021, Virtual Event, July 20--23, 2021, Proceedings, Part I 33}.\hskip 1em plus 0.5em minus 0.4em\relax Springer, 2021, pp. 3--25.

\bibitem{sun2022causality}
B.~Sun, J.~Sun, L.~H. Pham, and J.~Shi, ``Causality-based neural network repair,'' in \emph{Proceedings of the 44th International Conference on Software Engineering}, 2022, pp. 338--349.

\bibitem{ma2024vere}
J.~Ma, P.~Yang, J.~Wang, Y.~Sun, C.-C. Huang, and Z.~Wang, ``Vere: Verification guided synthesis for repairing deep neural networks,'' in \emph{Proceedings of the 46th IEEE/ACM International Conference on Software Engineering}, 2024, pp. 1--13.

\bibitem{chen2024isolation}
J.~Chen, J.~Wang, Y.~Sun, P.~Cheng, and J.~Chen, ``Isolation-based debugging for neural networks,'' in \emph{Proceedings of the 33rd ACM SIGSOFT International Symposium on Software Testing and Analysis}, 2024, pp. 338--349.

\bibitem{wang2024detoxifying}
M.~Wang, N.~Zhang, Z.~Xu, Z.~Xi, S.~Deng, Y.~Yao, Q.~Zhang, L.~Yang, J.~Wang, and H.~Chen, ``Detoxifying large language models via knowledge editing,'' in \emph{Proceedings of the 62nd Annual Meeting of the Association for Computational Linguistics (Volume 1: Long Papers)}, 2024, pp. 3093--3118.

\bibitem{imtiaz2025irepair}
S.~M. Imtiaz, A.~Singh, F.~Batole, and H.~Rajan, ``Irepair: An intent-aware approach to repair data-driven errors in large language models,'' \emph{Proceedings of the ACM on Software Engineering}, vol.~2, no. FSE, pp. 1226--1248, 2025.

\bibitem{havrilla2024understanding}
A.~Havrilla and M.~Iyer, ``Understanding the effect of noise in llm training data with algorithmic chains of thought,'' \emph{arXiv preprint arXiv:2402.04004}, 2024.

\bibitem{paul2021deep}
M.~Paul, S.~Ganguli, and G.~K. Dziugaite, ``Deep learning on a data diet: Finding important examples early in training,'' \emph{Advances in neural information processing systems}, vol.~34, pp. 20\,596--20\,607, 2021.

\bibitem{zheng2023coverage}
H.~Zheng, R.~Liu, F.~Lai, and A.~Prakash, ``Coverage-centric coreset selection for high pruning rates,'' in \emph{International Conference on Learning Representations (ICLR)}, 2023.

\bibitem{maharana2023d2}
A.~Maharana, P.~Yadav, and M.~Bansal, ``D2 pruning: Message passing for balancing diversity and difficulty in data pruning,'' \emph{arXiv preprint arXiv:2310.07931}, 2023.

\bibitem{zhangstaff}
X.~Zhang, J.~Zhai, S.~Ma, C.~Shen, T.~Li, W.~Jiang, and Y.~Liu, ``Staff: Speculative coreset selection for task-specific fine-tuning,'' in \emph{The Thirteenth International Conference on Learning Representations}, 2025.

\bibitem{hochbaum1985best}
D.~S. Hochbaum and D.~B. Shmoys, ``A best possible heuristic for the k-center problem,'' \emph{Mathematics of operations research}, vol.~10, no.~2, pp. 180--184, 1985.

\bibitem{nguyen2004active}
H.~T. Nguyen and A.~Smeulders, ``Active learning using pre-clustering,'' in \emph{Proceedings of the twenty-first international conference on Machine learning}, 2004, p.~79.

\bibitem{ertekin2007learning}
S.~Ertekin, J.~Huang, L.~Bottou, and L.~Giles, ``Learning on the border: active learning in imbalanced data classification,'' in \emph{Proceedings of the sixteenth ACM conference on Conference on information and knowledge management}, 2007, pp. 127--136.

\bibitem{sener2017active}
O.~Sener and S.~Savarese, ``Active learning for convolutional neural networks: A core-set approach,'' \emph{arXiv preprint arXiv:1708.00489}, 2017.

\bibitem{fawzi2016robustness}
A.~Fawzi, S.-M. Moosavi-Dezfooli, and P.~Frossard, ``Robustness of classifiers: from adversarial to random noise,'' \emph{Advances in neural information processing systems}, vol.~29, 2016.

\bibitem{merity2017pointer}
\BIBentryALTinterwordspacing
S.~Merity, C.~Xiong, J.~Bradbury, and R.~Socher, ``Pointer sentinel mixture models,'' in \emph{International Conference on Learning Representations}, 2017. [Online]. Available: \url{https://openreview.net/forum?id=Byj72udxe}
\BIBentrySTDinterwordspacing

\bibitem{paperno-etal-2016-lambada}
D.~Paperno, G.~Kruszewski, A.~Lazaridou, N.~Q. Pham, R.~Bernardi, S.~Pezzelle, M.~Baroni, G.~Boleda, and R.~Fern{\'a}ndez, ``The {LAMBADA} dataset: Word prediction requiring a broad discourse context,'' in \emph{Proceedings of the 54th Annual Meeting of the Association for Computational Linguistics (Volume 1: Long Papers)}, 2016, pp. 1525--1534.

\bibitem{gu2017badnets}
T.~Gu, B.~Dolan-Gavitt, and S.~Garg, ``Badnets: Identifying vulnerabilities in the machine learning model supply chain,'' \emph{arXiv preprint arXiv:1708.06733}, 2017.

\bibitem{szegedy2013intriguing}
C.~Szegedy, W.~Zaremba, I.~Sutskever, J.~Bruna, D.~Erhan, I.~Goodfellow, and R.~Fergus, ``Intriguing properties of neural networks,'' \emph{arXiv preprint arXiv:1312.6199}, 2013.

\bibitem{fu2022sound}
F.~Fu and W.~Li, ``Sound and complete neural network repair with minimality and locality guarantees,'' in \emph{International Conference on Learning Representations}, 2022.

\bibitem{hase2023does}
P.~Hase, M.~Bansal, B.~Kim, and A.~Ghandeharioun, ``Does localization inform editing? surprising differences in causality-based localization vs. knowledge editing in language models,'' \emph{Advances in Neural Information Processing Systems}, vol.~36, pp. 17\,643--17\,668, 2023.

\bibitem{yao2024survey}
Y.~Yao, J.~Duan, K.~Xu, Y.~Cai, Z.~Sun, and Y.~Zhang, ``A survey on large language model (llm) security and privacy: The good, the bad, and the ugly,'' \emph{High-Confidence Computing}, p. 100211, 2024.

\bibitem{weiser1984program}
M.~Weiser, ``Program slicing,'' \emph{IEEE Transactions on software engineering}, no.~4, pp. 352--357, 1984.

\bibitem{vessey1985expertise}
I.~Vessey, ``Expertise in debugging computer programs: A process analysis,'' \emph{International Journal of Man-Machine Studies}, vol.~23, no.~5, pp. 459--494, 1985.

\bibitem{liu2022improved}
Z.~Liu, Y.~Xu, Y.~Xu, Q.~Qian, H.~Li, X.~Ji, A.~Chan, and R.~Jin, ``Improved fine-tuning by better leveraging pre-training data,'' \emph{Advances in Neural Information Processing Systems}, vol.~35, pp. 32\,568--32\,581, 2022.

\bibitem{dettmers2023qlora}
T.~Dettmers, A.~Pagnoni, A.~Holtzman, and L.~Zettlemoyer, ``Qlora: Efficient finetuning of quantized llms,'' \emph{Advances in neural information processing systems}, vol.~36, pp. 10\,088--10\,115, 2023.

\bibitem{welling2009herding}
M.~Welling, ``Herding dynamical weights to learn,'' in \emph{Proceedings of the 26th annual international conference on machine learning}, 2009, pp. 1121--1128.

\bibitem{feldman2011scalable}
D.~Feldman, M.~Faulkner, and A.~Krause, ``Scalable training of mixture models via coresets,'' \emph{Advances in neural information processing systems}, vol.~24, 2011.

\bibitem{phillips2017coresets}
J.~M. Phillips, ``Coresets and sketches,'' in \emph{Handbook of Discrete and Computational Geometry}.\hskip 1em plus 0.5em minus 0.4em\relax Chapman and Hall/CRC, 2017, pp. 1269--1288.

\bibitem{mirzasoleiman2020coresets}
B.~Mirzasoleiman, J.~Bilmes, and J.~Leskovec, ``Coresets for data-efficient training of machine learning models,'' in \emph{International Conference on Machine Learning}.\hskip 1em plus 0.5em minus 0.4em\relax PMLR, 2020, pp. 6950--6960.

\bibitem{albalak2024survey}
A.~Albalak, Y.~Elazar, S.~M. Xie, S.~Longpre, N.~Lambert, X.~Wang, N.~Muennighoff, B.~Hou, L.~Pan, H.~Jeong \emph{et~al.}, ``A survey on data selection for language models,'' \emph{arXiv preprint arXiv:2402.16827}, 2024.

\bibitem{lin2024data}
X.~Lin, W.~Wang, Y.~Li, S.~Yang, F.~Feng, Y.~Wei, and T.-S. Chua, ``Data-efficient fine-tuning for llm-based recommendation,'' in \emph{Proceedings of the 47th international ACM SIGIR conference on research and development in information retrieval}, 2024, pp. 365--374.

\bibitem{wang2023farewell}
X.~Wang, W.~Zhou, Q.~Zhang, J.~Zhou, S.~Gao, J.~Wang, M.~Zhang, X.~Gao, Y.~W. Chen, and T.~Gui, ``Farewell to aimless large-scale pretraining: Influential subset selection for language model,'' in \emph{Findings of the Association for Computational Linguistics: ACL 2023}, 2023, pp. 555--568.

\bibitem{joaquin2024in2core}
A.~Joaquin, B.~Wang, Z.~Liu, P.~Muller, N.~Asher, B.~Lim, and N.~Chen, ``In2core: Leveraging influence functions for coreset selection in instruction finetuning of large language models,'' in \emph{Findings of the Association for Computational Linguistics: EMNLP 2024}, 2024, pp. 10\,324--10\,335.

\bibitem{xia2024less}
M.~Xia, S.~Malladi, S.~Gururangan, S.~Arora, and D.~Chen, ``Less: Selecting influential data for targeted instruction tuning,'' in \emph{International Conference on Machine Learning}.\hskip 1em plus 0.5em minus 0.4em\relax PMLR, 2024, pp. 54\,104--54\,132.

\bibitem{chan2022redunet}
K.~H.~R. Chan, Y.~Yu, C.~You, H.~Qi, J.~Wright, and Y.~Ma, ``Redunet: A white-box deep network from the principle of maximizing rate reduction,'' \emph{Journal of machine learning research}, vol.~23, no. 114, pp. 1--103, 2022.

\bibitem{seki2024diversity}
K.~Seki, S.~Takamichi, T.~Saeki, and H.~Saruwatari, ``Diversity-based core-set selection for text-to-speech with linguistic and acoustic features,'' in \emph{ICASSP 2024-2024 IEEE International Conference on Acoustics, Speech and Signal Processing (ICASSP)}.\hskip 1em plus 0.5em minus 0.4em\relax IEEE, 2024, pp. 12\,351--12\,355.

\bibitem{radford2019language}
A.~Radford, J.~Wu, R.~Child, D.~Luan, D.~Amodei, I.~Sutskever \emph{et~al.}, ``Language models are unsupervised multitask learners,'' \emph{OpenAI blog}, vol.~1, no.~8, p.~9, 2019.

\bibitem{gao2020pile}
L.~Gao, S.~Biderman, S.~Black, L.~Golding, T.~Hoppe, C.~Foster, J.~Phang, H.~He, A.~Thite, N.~Nabeshima \emph{et~al.}, ``The pile: An 800gb dataset of diverse text for language modeling,'' \emph{arXiv preprint arXiv:2101.00027}, 2020.

\bibitem{biderman2023pythia}
S.~Biderman, H.~Schoelkopf, Q.~G. Anthony, H.~Bradley, K.~O’Brien, E.~Hallahan, M.~A. Khan, S.~Purohit, U.~S. Prashanth, E.~Raff \emph{et~al.}, ``Pythia: A suite for analyzing large language models across training and scaling,'' in \emph{International Conference on Machine Learning}.\hskip 1em plus 0.5em minus 0.4em\relax PMLR, 2023, pp. 2397--2430.

\bibitem{lee2024mechanistic}
A.~Lee, X.~Bai, I.~Pres, M.~Wattenberg, J.~K. Kummerfeld, and R.~Mihalcea, ``A mechanistic understanding of alignment algorithms: A case study on dpo and toxicity,'' in \emph{International Conference on Machine Learning}.\hskip 1em plus 0.5em minus 0.4em\relax PMLR, 2024, pp. 26\,361--26\,378.

\bibitem{machalica2019predictive}
M.~Machalica, A.~Samylkin, M.~Porth, and S.~Chandra, ``Predictive test selection,'' in \emph{2019 IEEE/ACM 41st International Conference on Software Engineering: Software Engineering in Practice (ICSE-SEIP)}.\hskip 1em plus 0.5em minus 0.4em\relax IEEE, 2019, pp. 91--100.

\bibitem{chen2024fast}
J.~Chen, J.~Wang, X.~Zhang, Y.~Sun, M.~Kwiatkowska, J.~Chen, and P.~Cheng, ``Fast: Boosting uncertainty-based test prioritization methods for neural networks via feature selection,'' in \emph{Proceedings of the 39th IEEE/ACM International Conference on Automated Software Engineering}, 2024, pp. 895--906.

\bibitem{liu2023chain}
H.~Liu, C.~Sferrazza, and P.~Abbeel, ``Chain of hindsight aligns language models with feedback,'' \emph{arXiv preprint arXiv:2302.02676}, 2023.

\bibitem{perspectiveapi2025}
\BIBentryALTinterwordspacing
``Perspective api,'' 2025, retrieved July 10, 2025. [Online]. Available: \url{https://perspectiveapi.com/}
\BIBentrySTDinterwordspacing

\bibitem{geva2022Transformer}
M.~Geva, A.~Caciularu, K.~Wang, and Y.~Goldberg, ``Transformer feed-forward layers build predictions by promoting concepts in the vocabulary space,'' in \emph{Proceedings of the 2022 Conference on Empirical Methods in Natural Language Processing}, 01 2022, pp. 30--45.

\end{thebibliography}

\newpage

 




\vfill

\end{document}


\appendices
\section{Implementation Details}\label{appendix_a}
\subsection{Embedding Extraction}
We adopt the all-MiniLM-L6-v2 encoder \cite{huggingface-allminilm} to extract embeddings for all samples. Each input $x_i$ is mapped into a $d$-dimensional representation $v_i \in \mathbb{R}^d$, forming the embedding matrix
\begin{equation}
V = [v_1, v_2, \ldots, v_N]^{\top} \in \mathbb{R}^{N \times d}.
\end{equation}
This encoder is selected because it offers a favorable trade-off between computational efficiency and semantic richness, making it particularly well-suited for large-scale experimental settings where both accuracy and speed are critical. Its ability to capture fine-grained semantic similarity across diverse textual inputs ensures that downstream clustering is based on meaningful structural information rather than superficial lexical overlap.

\subsection{Dimensionality Reduction}
To mitigate redundancy and noise in the high-dimensional embedding space, we apply principal component analysis (PCA) \cite{abdi2010principal}. The number of retained dimensions is determined by analyzing the trend of cumulative explained variance, allowing us to preserve the majority of semantic information while discarding less informative components. Following this criterion, we reduce the dimensionality to $d'=50$, obtaining
\begin{equation}
\hat{v}_i = \mathbf{PCA}(v_i), \quad 
\hat{V} = [\hat{v}_1, \hat{v}_2, \ldots, \hat{v}_N]^{\top} \in \mathbb{R}^{N \times d'}.
\end{equation}
This data-driven selection of dimensionality avoids arbitrary thresholds and ensures that the reduced space maintains a balance between compactness and representational fidelity, which is essential for reliable clustering in the subsequent stage.

\subsection{Clustering}
On the reduced embeddings $\hat{V}$, we employ the $k$-means clustering algorithm \cite{macqueen1967some} to partition the samples. The number of clusters is set to $K=10$, reflecting the semantic composition of the dataset. Formally,
\begin{equation}
\mathcal{C} = \mathbf{KMeans}(\hat{V}, K), \qquad
\mathcal{C} = \{\mathcal{C}_1, \mathcal{C}_2, \ldots, \mathcal{C}_K\}.
\end{equation}
The choice of $k$-means is motivated by its efficiency, scalability, and inductive bias toward discovering spherical clusters, which aligns well with the observed distribution of language model embeddings. Fixing the number of clusters according to dataset semantics further enables stable and interpretable grouping, ensuring that representative subsets are identified consistently across repair pipelines.